\PassOptionsToPackage{table}{xcolor}

\documentclass[10pt,twocolumn,letterpaper]{article}

\usepackage{cvpr}              % To produce the CAMERA-READY version
\usepackage[accsupp]{axessibility}  % Improves PDF readability for those with disabilities.

\usepackage{microtype}

\definecolor{cvprblue}{rgb}{0.21,0.49,0.74}
\usepackage[breaklinks,colorlinks,allcolors=cvprblue]{hyperref}
\usepackage{multirow}
\usepackage{colortbl}
\usepackage{booktabs,array}
\newcolumntype{C}[1]{>{\centering\arraybackslash}p{#1}}

\newcolumntype{L}[1]{>{\raggedright\arraybackslash}p{#1}}
\newcolumntype{R}[1]{>{\raggedleft\arraybackslash}p{#1}}
\newcommand{\compacttableset}{%
  \scriptsize% 
  \setlength{\tabcolsep}{4pt}% 
  \renewcommand{\arraystretch}{1.05}%
}
\usepackage[table]{xcolor}
\definecolor{brightgreen}{RGB}{0,150,70}
\usepackage{pifont}
\usepackage{CJKutf8}
\usepackage{marvosym}

\newenvironment{compactmath}{%
  \begingroup
  \small
  \setlength{\abovedisplayskip}{4pt}%
  \setlength{\belowdisplayskip}{4pt}%
  \setlength{\abovedisplayshortskip}{2pt}%
  \setlength{\belowdisplayshortskip}{2pt}%
  \setlength{\jot}{2pt}%
}{\endgroup}

\newif\ifshowcomments
\showcommentsfalse   % ← hide comments

\definecolor{lowyellow}{RGB}{241, 196, 15}

\usepackage{tikz}
\usepackage{graphicx}
\def\paperID{31878} % *** Enter the Paper ID here
\def\confName{CVPR}
\def\confYear{2026}

\title{LongVT: Incentivizing ``Thinking with Long Videos'' via Native Tool Calling}

\author{Zuhao Yang$^{1,2,5,*}$, Sudong Wang$^{1,3,5,*}$, Kaichen Zhang$^{1,2,5,*}$, Keming Wu$^{1,4,5}$, Sicong Leng$^{2}$, \\
Yifan Zhang$^{1}$, Bo Li$^{2,5}$, Chengwei Qin$^{3}$, Shijian Lu$^{2,\text{\Letter}}$, Xingxuan Li$^{1,\text{\Letter}}$, Lidong Bing$^{1}$\\
$^{1}$MiroMind, $^{2}$NTU, $^{3}$HKUST(GZ), $^{4}$THU, $^{5}$LMMs-Lab \\
}

\AddToHook{begindocument/before}{%
  \expandafter\let\csname \string\showhyphens\endcsname\showhyphens
}

\begin{document}

\twocolumn[{%
   \renewcommand\twocolumn[1][]{#1}%
   \maketitle
   \vspace{-30pt}
   \begin{center}
    \centering
    \centerline{\small \url{https://evolvinglmms-lab.github.io/LongVT/}}
    \vspace{1em}
    \includegraphics[width=0.95\textwidth]{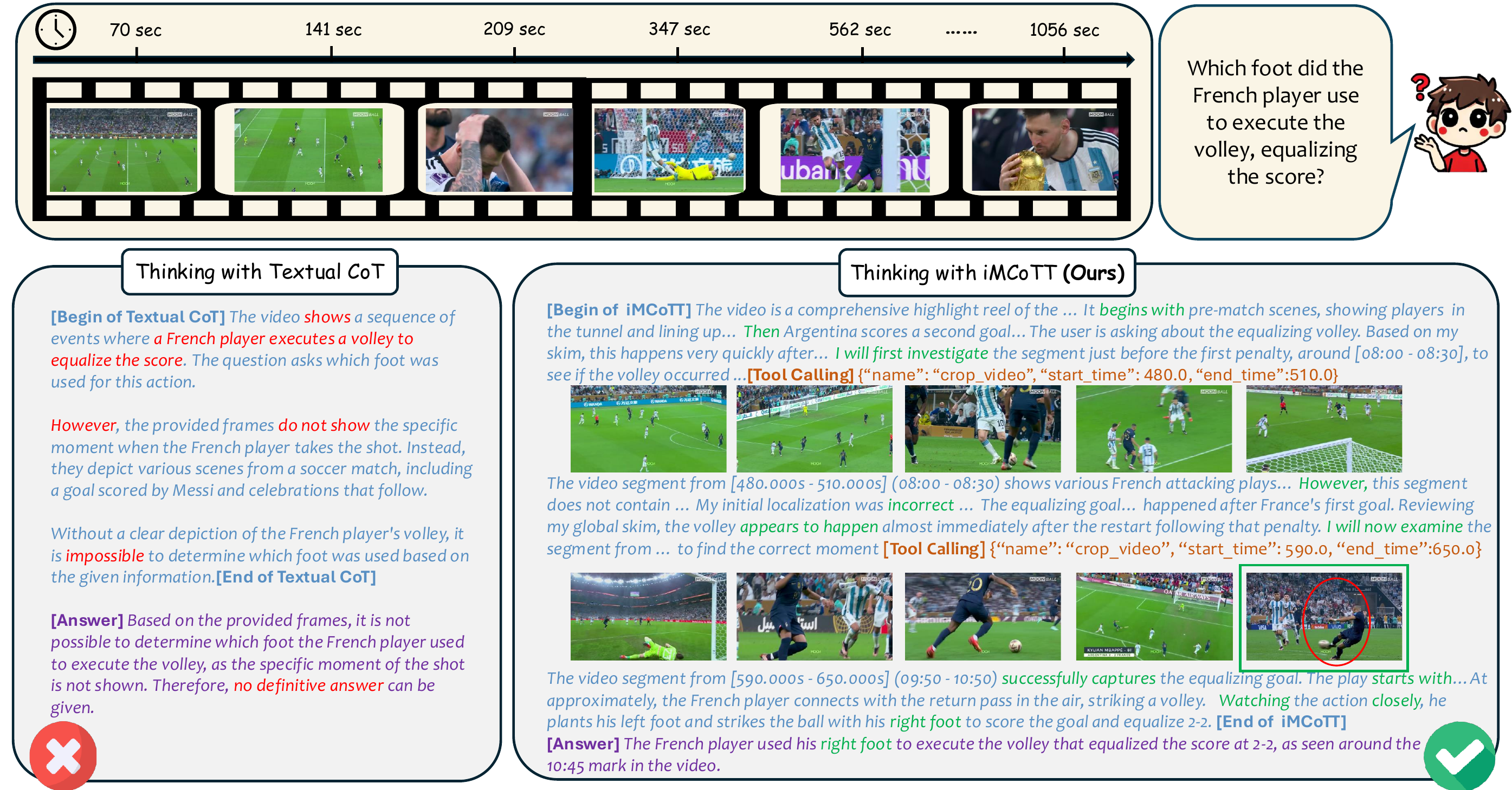}
    \captionsetup{hypcap=false} 
    \captionof{figure}{\textbf{Interleaved Multimodal Chain-of-Tool-Thought (iMCoTT).} 
    Compared to prior text-based Chain-of-Thought (CoT) reasoning, iMCoTT in our proposed \textit{\textbf{LongVT}} can \emph{natively} perform self-reflection via \emph{calling} \texttt{crop\_video(start\_time, end\_time)} \emph{tool}.
    It \textcolor{brightgreen}{proposes} a \textcolor{brightgreen}{time window} after a global preview, proactively \textcolor{brightgreen}{fetches} the corresponding \textcolor{brightgreen}{short clip}, \textcolor{brightgreen}{rethinks} based on the \textcolor{brightgreen}{new evidence}, and \textcolor{brightgreen}{determines} whether to \textcolor{brightgreen}{refine} or \textcolor{brightgreen}{answer directly}.
    Such tool-augmented reasoning behaviors ground each step in what is actually seen rather than \textcolor{red}{blindly rephrasing} in text-only CoT, which mitigates \textcolor{red}{hallucination} and leads to \textcolor{brightgreen}{enhanced temporal localization} and \textcolor{brightgreen}{answer correctness}.}
    \label{fig:teaser}
   \end{center}%
}]

\renewcommand{\thefootnote}{\fnsymbol{footnote}}
\footnotetext[1]{\raggedright Equal Contribution \quad \textsuperscript{\Letter}Corresponding Authors: \texttt{Shijian.Lu@ntu.edu.sg}, \texttt{xingxuan.li@miromind.ai}}
\renewcommand*{\thefootnote}{\arabic{footnote}}

\begin{abstract}
Large multimodal models (LMMs) have shown great potential for video reasoning with textual Chain-of-Thought.
However, they remain vulnerable to hallucinations, especially when processing long-form videos where evidence is sparse and temporally dispersed.
Inspired by how humans comprehend long videos\textemdash{}by first skimming globally and then examining relevant clips for details\textemdash{}we introduce \textbf{LongVT}, an end-to-end agentic framework that enables ``Thinking with \textbf{Long} \textbf{V}ideos'' via interleaved Multimodal Chain-of-\textbf{T}ool-Thought.
Specifically, we exploit LMMs' inherent temporal grounding ability as a native video cropping tool to zoom in on a specific video clip and resample finer-grained video frames.
This global-to-local reasoning loop continues until answers are grounded in retrieved visual evidence.
Given the scarcity of fine-grained question-answering (QA) data for the long video reasoning task, we curate and release a data suite named \textbf{VideoSIAH} to facilitate both training and evaluation.
Specifically, our training dataset consists of 247.9K samples for tool-integrated cold-start supervised fine-tuning, 1.6K samples for agentic reinforcement learning, and 15.4K samples for agentic reinforcement fine-tuning.
Our evaluation benchmark consists of 652 QA pairs that are carefully curated through a semi-automatic data pipeline with human-in-the-loop validation.
With a meticulously designed three-stage training strategy and extensive empirical validation, LongVT consistently outperforms existing strong baselines across four challenging long-video understanding and reasoning benchmarks.
Our code, data, and model weights are publicly available at \url{https://github.com/EvolvingLMMs-Lab/LongVT}.
\end{abstract}    
\section{Introduction}
\label{sec:intro}

Understanding long-form videos ($>$15 minutes) poses a major challenge in multimodal intelligence \cite{fu2025videomme,hu2025videommmu,wu2024longvideobench,wang2025lvbench}.
Compared with short clips, long videos contain complex event structures and require sustained comprehension across thousands of frames, supporting tasks such as video question answering (QA) \cite{li2024mvbench,liu2024tempcompass,cai2024temporalbench,wu2024longvideobench,wang2025lvbench}, temporal grounding \cite{gao2017charades,krishna2017anetcaptions,yang2023vidchapters,ren2024timechat,yang2025timeexpert}, and dense captioning \cite{zhou2018youcook2,krishna2017anetcaptions,huang2020vitt}.
These capabilities underpin applications such as soccer event spotting \cite{li2025f16} and film understanding \cite{song2024moviechat}.
Recent LMMs \cite{feng2025videor1,li2025videochatr1,wang2025videorft,wang2025timer1,zhang2025videollama3} exhibit promising short video reasoning, yet most rely on the R1-style paradigm \cite{guo2025deepseekr1}\textemdash{}supervised fine-tuning (SFT) with textual Chain-of-Thought (CoT), followed by Group Relative Policy Optimization (GRPO)-based reinforcement learning (RL) \cite{shao2024grpo}.
Such pipelines remain largely language-centric, limiting visual reasoning and increasing hallucinations in long-video scenarios \cite{zhang2025vital}.
Moreover, their uniform sampling fails to adaptively capture key visual evidence, often missing fine-grained or decisive moments critical for long-video reasoning.
This motivates our central question: \emph{Can LMMs reliably reason over long videos by performing human-like visual operations to guide their reasoning?}

Let us consider the following scenario: a testee is asked to answer the question, ``Which foot did the French player use to execute the volley, equalizing the score?'' using only the silent video of a football match. Without audio, metadata, or timeline markers, the testee must rely purely on visual inspection. Based on common viewing habits, a human would typically jump through the video in coarse intervals, searching for strong visual indicators of a goal—such as crowd reactions, player celebrations, referee gestures, or scoreboard updates. After locating a likely scoring segment, the testee would rewind slightly and examine the surrounding frames more carefully to pinpoint the exact equalizing moment, and then verify the scoring foot using close-up shots. Notably, when we prompt two state-of-the-art proprietary LMMs (i.e., GPT-5 \cite{openai2025gpt5} and Gemini 2.5 Pro \cite{comanici2025gemini25}) with the same task, the strategies they propose closely mirror this human-intuitive procedure (see the Supplementary Material).

As illustrated in \Cref{fig:teaser}, the testee, seeking to save time, avoids scanning the entire video frame by frame. Instead, they first perform a coarse global skim and then zoom in on promising segments. When projected to the LMM setting, this global-to-local reasoning strategy enables models with limited context length to process extremely long videos effectively.
To implement such a strategy, we design interleaved Multimodal Chain-of-Tool-Thought (iMCoTT) that enables LMMs to naturally interleave reasoning with on-demand temporal retrieval via dynamically selecting and re-inspecting video segments of interest.
Such LMM behaviors stem from \emph{latent} temporal grounding capabilities that are activated through tool-integrated fine-tuning, without an auxiliary expert model or external retriever.
Our designed iMCoTT enables ``looking again'' by proposing a more robust time window, examining that snippet, and revising its hypothesis when necessary.
This helps reduce hallucinations and reveals finer details, akin to human self-reflection upon realizing an initially inspected segment was erroneous.

This human-inspired ``Thinking with Long Videos'' paradigm is naturally suitable for queries that either require aggregating clues across multiple shots or hinge on a brief and evidence-bearing segment within hours-long footage.
Yet, the open-source community lacks training and evaluation data with such fine-grained queries: most public datasets emphasize general and high-level questions but rarely train and evaluate reasoning capability under a ``\textbf{Video} \textbf{S}egment-\textbf{I}n-\textbf{A}-\textbf{H}aystack'' setting.
We address this grand challenge by constructing \textbf{VideoSIAH} that comprises high-quality QA pairs and tool-augmented reasoning traces.
VideoSIAH comprises 247.9K samples for SFT, 1.6K samples for agentic RL, and 15.4K samples for reinforcement fine-tuning (RFT). 
Besides, we curate a dedicated evaluation benchmark, \textbf{VideoSIAH-Eval}, comprising 652 QA pairs that have undergone human-in-the-loop validation \cite{cakmak2014humanintheloop}, where each question's supporting evidence lies within a narrow window relative to the full video duration.

In this paper, we introduce \textbf{LongVT}, an end-to-end agentic framework that elicits LMMs' ability for ``Thinking with \textbf{Long V}ideos'' via a three-stage training strategy with large-scale and high-quality \textbf{T}ool-augmented data from VideoSIAH.
The \emph{first} stage performs cold-start SFT that empowers the base LMM with three fundamental capabilities: \textbf{(1)} proposing a precise window for relevant event(s), \textbf{(2)} reasoning over densely resampled frames within the window, and \textbf{(3)} self-correcting when the window is suboptimal.
The \emph{second} stage adopts agentic RL for enhancing the model's generalization over open-ended QA tasks.
Unlike existing work that relies on answer-only rewards for video QA and IoU rewards for temporal grounding \cite{feng2025videor1, wang2025timer1}, we design a joint answer-temporal grounding reward function that explicitly encourages exploratory rollouts with improved temporal localization while preserving answer correctness.
The \emph{third} stage leverages agentic RFT where the model is further optimized by utilizing filtered rollout traces distilled from its own RL-trained policy. 
This stage stabilizes agentic behaviors learned during RL and consolidates fine-grained temporal localization and multi-step reasoning. 

The contributions of our work can be summarized in three major aspects.
\textbf{First}, we introduce an end-to-end agentic paradigm that natively interleaves multimodal tool-augmented CoT with on-demand clip inspection over hours-long videos, thereby enabling LMMs to perform more effective and reliable long-video reasoning.
\textbf{Second}, to facilitate training and evaluation of evidence-sparse long-video reasoning, we construct a scalable data pipeline that produces diverse and high-quality QAs and tool-integrated reasoning traces, and a dedicated benchmark under a video segment-in-a-haystack setting.
\textbf{Third}, we conduct comprehensive ablations and are the first to systematically validate that the joint answer-temporal grounding reward and RFT each yield consistent gains over SFT and RL alone for long-video reasoning, establishing a state-of-the-art baseline with invaluable insights.

\section{Related Work}
\label{sec:related}

\paragraph{RL-Based Multimodal Reasoning.}
Inspired by large reasoning models such as OpenAI o1 \cite{openai2024o1} and DeepSeek-R1 \cite{guo2025deepseekr1}, recent studies extend GRPO-style RL from text-only reasoning to multimodal domains. In vision, methods enhance reasoning for image QA \cite{huang2025visionr1,meng2025mmeureka,leng2025mmr1,zhang2025openmmreasoner}, grounding \cite{liu2025visualrft,shen2025vlmr1,fan2025grit}, and segmentation \cite{liu2025segzero}. Video-centric approaches further tackle temporal reasoning tasks such as video QA \cite{feng2025videor1,wang2025videorft}, temporal grounding \cite{wang2025timer1}, and spatio-temporal grounding \cite{li2025videochatr1}, including recent efforts to scale RL to long videos \cite{chen2025longrl,tian2025egor1}. Audio-aware methods similarly apply RL to audio QA \cite{li2025r1aqa,wen2025sari} and broader omnimodal reasoning \cite{zhong2025omnir1}. Complementary efforts target the SFT-to-RL transition itself, with pre-alignment distillation \cite{wang2026prism} stabilizing the RL starting point. Collectively, these works demonstrate that RL-based reasoning improves cross-modal understanding.

\paragraph{Tool-Augmented Agentic LMMs.}
Complementing RL-based reasoning, another line of research incorporates tools to incentivize LMMs' agentic capabilities. For images, recent methods \cite{zheng2025deepeyes,su2025pixelreasoner,wu2025vilasr,yang2025mermaid} interleave pixel-level operations (e.g., zooming in, drawing auxiliary lines, generative imagery) to reason over finer details while reducing hallucinations. For videos, VITAL \cite{zhang2025vital} shows that tool-augmented RL improves video QA and temporal grounding, while concurrent multi-agent designs target parallel tool dispatch \cite{yang2026paravt} or storyline-guided cross-modal collaboration \cite{yang2026svagent}. \emph{Our approach differs from VITAL in two key aspects.}
\textbf{First}, LongVT targets video segment-in-a-haystack reasoning and contributes a large-scale, high-quality dataset and benchmark. VideoSIAH not only triggers tool-integrated reasoning but also reveals emergent human-like self-reflection in long-form video understanding.
\textbf{Second}, we propose a three-stage closed-loop training paradigm combining SFT cold start, RL, and a dedicated RFT stage leveraging high-quality rollout traces for iterative self-refinement.
\textbf{Moreover}, unlike prior work relying on multi-task objectives \cite{feng2025videor1,li2025videochatr1} or explicit tool rewards \cite{zheng2025deepeyes,zhang2025vital}, LongVT shows that single-task RL with a decoupled temporal-grounding reward can still achieve state-of-the-art performance in long video reasoning.

\section{VideoSIAH: A Fine-Grained Data Suite for Evidence-Sparse Long-Video Reasoning}
\label{sec:data}

Long-video reasoning presents a fundamentally different challenge from previous video QA settings: 
LMMs must locate \emph{sparse, fine-grained, and causally decisive} moments embedded within hours-long content. 
However, existing tool-augmented LMMs \cite{su2025pixelreasoner,zhang2025vital} are mostly trained with \emph{coarse-grained and clip-level} data.
This mismatch leaves modern LMMs lacking the supervision needed to learn how temporal hypotheses are formed, verified, or revised\textemdash{}a critical yet underexplored capability for agentic long-video reasoning.
Moreover, most existing video understanding benchmarks \cite{fu2025videomme,wu2024longvideobench,wang2025lvbench} only offer multiple-choice QAs, which can be solved without genuine temporal grounding and are vulnerable to dataset leakage or shortcut exploitation. Evidence and discussion can be found in the Supplementary Material.
To fill this gap, we introduce \textbf{VideoSIAH}, a large-scale, diverse, and high-quality data suite that serves collectively as a training dataset capturing the reasoning dynamics required for segment-in-a-haystack question-answering, and a fine-grained evaluation benchmark, \textbf{VideoSIAH-Eval}, with human-in-the-loop validation for long-video open-ended question-answering. 

\begin{figure*}[t]
    \centering
    \includegraphics[width=\linewidth]{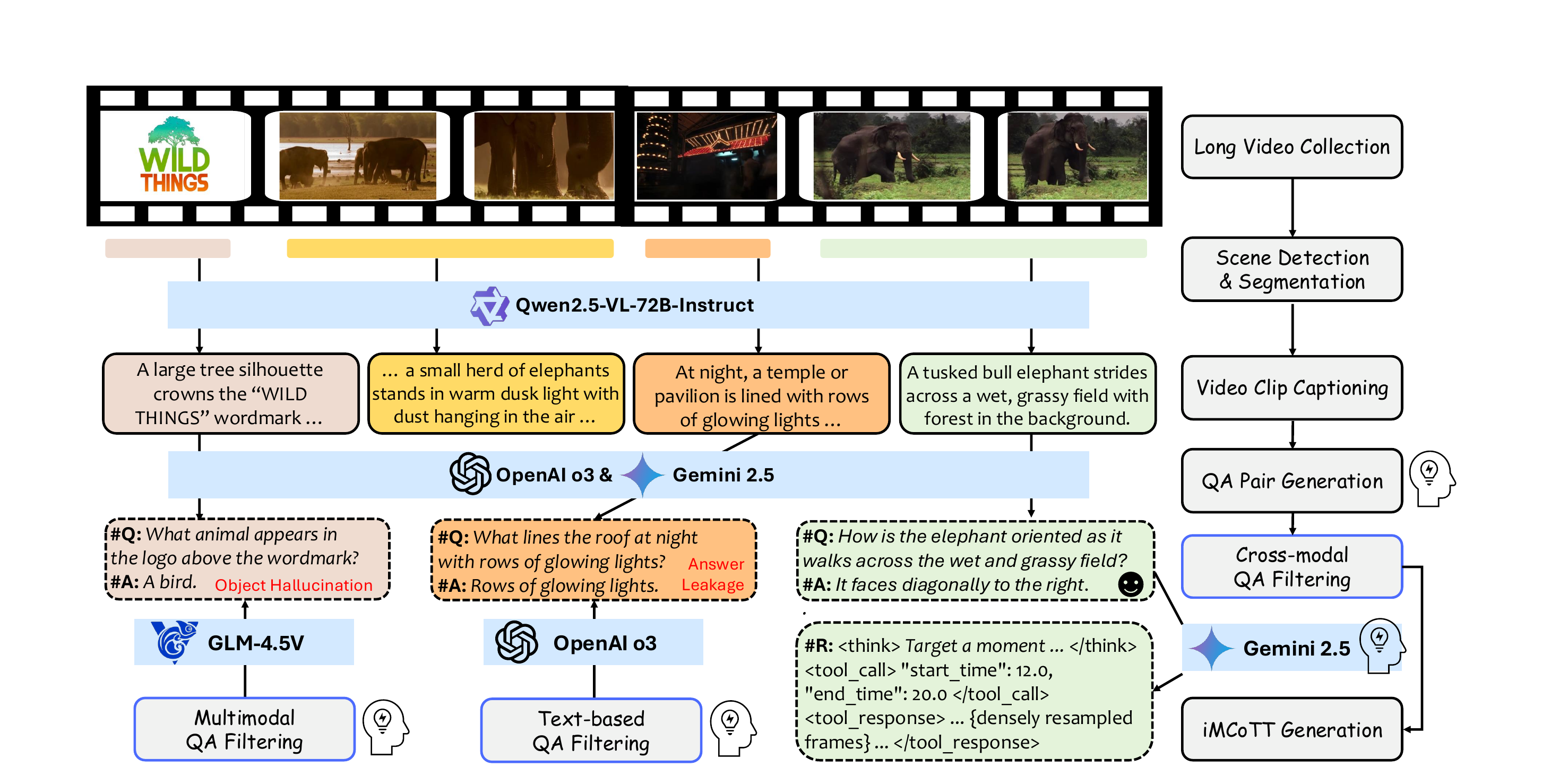}
    \caption{\textbf{Data Pipeline of \textbf{VideoSIAH}.}
    We construct a semi-automatic data pipeline that integrates several state-of-the-art LMMs \cite{bai2025qwen25vl,openai2025o3,comanici2025gemini25,hong2025glm45v} to sequentially perform long video segmentation, video clip captioning, segment-in-a-haystack QA generation, cross-modal QA filtering, and iMCoTT generation.
    Icons with human silhouettes denote human-in-the-loop validation, where annotators inspect a small set of representative failures to refine prompting rules for QA generation, QA filtering, and iMCoTT generation.
    Note that iMCoTT traces are generated only for the cold-start SFT stage, 
    whereas RL training operates solely on the filtered QA pairs.}
    \label{fig:data_pipe}
\end{figure*}

\subsection{Data Pipeline}
\label{sec:data_pipeline}

As illustrated in \Cref{fig:data_pipe}, VideoSIAH is curated through a semi-automatic, human-in-the-loop pipeline that constructs temporally grounded reasoning traces aligned with human cognitive processes during evidence-sparse long-video reasoning.
We apply a deterministic, pixel-level scene detection algorithm on long videos and merge consecutive segments shorter than 10 seconds to obtain semantically stable units, ensuring that tool usage is grounded in visually coherent temporal intervals rather than random splits. For each segment, Qwen2.5-VL-72B \cite{bai2025qwen25vl} generates detailed descriptions capturing salient objects, spatial relations, and evolving events. These captions serve as the semantic basis for generating temporally grounded QA pairs. Initial QAs are created from the captions, covering temporal events, spatial layouts, motion, object attributes, and scene transitions, ensuring broad coverage at scale.

To ensure quality, we employ two filtering stages:
\textbf{(1)} text-based QA filtering, which removes low-quality or ill-posed QAs (e.g., answer leakage) using linguistic heuristics and model agreement; and
\textbf{(2)} multimodal QA filtering, where GLM-4.5V \cite{hong2025glm45v} verifies answer consistency against the video segment, eliminating hallucinated and visually unsupported claims.
Annotator feedback further refines prompting rules for QA generation, filtering, and iMCoTT construction. This prompt-feedback refinement loop boosts reliability without heavy manual annotation, yielding high-fidelity, temporally grounded, and scalable data.

\subsection{Dataset Curation}

\paragraph{SFT Data Curation.}
Our SFT data is constructed from three major categories: \textbf{(1)} tool-augmented multi-round data, \textbf{(2)} image reasoning data, and \textbf{(3)} video reasoning data, with the goal of enhancing both tool-calling capability and general reasoning performance. We curate tool-augmented QA pairs following the pipeline illustrated in \Cref{fig:data_pipe}. When processing hours-long videos, we find that sparsely sampled frames from a single round often fail to capture the correct temporal segment, which makes multi-round tool-calling necessary.
To address this limitation, we generate multi-round tool-calling traces in an adaptive manner based on video length. Specifically, we define the probability of selecting a sample for multi-round curation as
\begin{small}
$$
\label{eq:multi-round}
\begin{aligned}
P_{\text{multi}} = 1 - \frac{L_{\max} - \operatorname{clip}(L_{\text{video}}, L_{\min}, L_{\max})}{L_{\max} - L_{\min}},
\end{aligned}
$$
\end{small}
where \( P_{\text{multi}} \) denotes the probability of choosing a given data sample for multi-round generation, \( L_{\text{video}} \) represents the video length, and \( L_{\max} \) and \( L_{\min} \) are the maximum and minimum video length thresholds, respectively. The function \(\operatorname{clip}(x, a, b)\) restricts \(x\) to the range \([a, b]\). Videos selected under this criterion undergo multi-round data generation to ensure that longer videos receive proportionally more tool-calling rounds, improving temporal coverage and reasoning completeness. We further gather a mixture of diverse video and image reasoning datasets. 

\begin{table}[t]
\centering
\compacttableset
\resizebox{\columnwidth}{!}{%
\begin{tabular}{@{} C{0.19\linewidth} L{0.24\linewidth} L{0.34\linewidth} R{0.10\linewidth} C{0.10\linewidth} @{}}
\toprule[1pt]
\textbf{Split} & \textbf{Source} & \textbf{Purpose} & \textbf{Samples} & \textbf{Total} \\
\midrule\midrule
\multirow{3}{*}[-1.0em]{\textbf{SFT (w/o tool)}}
  & LongVideo-Reason CoT \cite{chen2025longrl} & Reasoning-augmented Open-ended QA & 5{,}238   & \multirow{3}{*}[-1.0em]{\textbf{228{,}835}} \\
  & Video-R1 CoT \cite{feng2025videor1}        & Reasoning-augmented Video QA      & 165{,}575 &  \\
  & Image-based CoT                            & Reasoning-augmented Image QA      & 58{,}022  &  \\
\cmidrule(lr){2-4}

\multirow{2}{*}{\textbf{SFT (w/ tool)}}
  & Gemini-distilled iMCoTT       & Tool-augmented Open-ended QA      & 12{,}766 & \multirow{3}{*}{\textbf{19{,}161}} \\
  & Qwen-distilled iMCoTT         & Tool-augmented Temporal Grounding & 6{,}395  &  \\
\cmidrule(lr){2-4}

\textbf{RL}
  & Gemini-distilled QAs          & Open-ended QA over Long Videos    & 1{,}667  & \multirow{2}{*}{\textbf{17{,}020}} \\
\textbf{RFT}
  & Self-distilled iMCoTT         & Agentic Behaviors                 & 15{,}353 &  \\
\bottomrule[1pt]
\end{tabular}%
}%
\caption{\textbf{Dataset Statistics of VideoSIAH.} Our proposed dataset contains non-tool SFT data, tool-augmented SFT data, RL QAs, and self-distilled RFT traces.}
\label{tab:data_dist}
\end{table}

\paragraph{RL Data Curation.} For RL, the split is built from the filtered segment-in-a-haystack QA pairs produced by our data pipeline in \Cref{sec:data_pipeline}. Each QA is associated with the length of its source video, and we partition candidates into several duration bands (short, medium, long). From these bands, we sample a length-balanced subset, ensuring the RL data is not dominated by very short clips and instead covers a diverse range of video durations. On top of this length-balanced pool, we apply a simple difficulty-aware filter based on multi-turn tool runs. For each question, we draw \(K\) rollouts of the current policy; if all \(K\) trajectories answer correctly (too easy) or all \(K\) fail (too hard), we discard the item and retain only questions with mixed outcomes. This focuses RL on a middle band of difficulty and avoids degenerate reward signals, yielding a more informative and stable optimization process.

\paragraph{RFT Data Curation.}
To construct the RFT traces, we filter trajectories from early RL runs and retain only high-quality cases. Concretely, a trajectory is kept if the model produces the correct final answer and its predicted temporal span attains an Intersection over Union (IoU) of at least 0.3 with the annotated ground-truth window.
This dual criterion enforces both semantic correctness and sufficiently accurate temporal grounding, ensuring the curated traces reflect genuinely successful long-video reasoning rather than reward hacking or lucky guesses. We then convert these filtered trajectories into supervised training examples for post-RL refinement. Training on this self-generated, well-grounded subset provides high-precision in-distribution supervision, stabilizes optimization, and further strengthens the model's grounding and tool-calling behavior beyond what SFT alone can provide.

\subsection{Dataset Statistics}
As shown in \Cref{tab:data_dist}, VideoSIAH comprises 228,835 SFT samples with normal (non-tool) CoT annotation, 19,161 tool-augmented SFT samples, and 17,020 instances used for RL and RFT.
In the SFT split, the non-tool portion is dominated by long-video reasoning data \cite{chen2025longrl}, complemented by Video-R1-CoT \cite{feng2025videor1} and a smaller amount of hard image-based CoT supervision. A detailed breakdown can be found in the Supplementary Material.
The tool-augmented subset combines Gemini 2.5 Flash \cite{comanici2025gemini25} distilled CoT traces (i.e., iMCoTT) for open-ended QA and Qwen2.5-VL-72B-Instruct \cite{bai2025qwen25vl} distilled traces for temporal grounding, providing joint supervision for tool usage and timestamp prediction.
For the RL split, we filtered a high-quality subset of QA instances from \Cref{sec:data_pipeline}. For RFT, we further select high-quality RL rollout traces for post-RL refinement, providing dense supervision that enables the policy to go well beyond the SFT-only performance ceiling.
Together, these components yield a large-scale and diverse dataset spanning SFT, RL, and RFT, covering high-level reasoning, temporal grounding, and tool-integrated behaviors.
For evaluation, we introduce the VideoSIAH-Eval benchmark, which consists of 244 videos and 652 carefully filtered QA pairs\footnote{An earlier release contained 1,280 entries due to unintentional duplication during data export; the cleaned version with 652 unique QA pairs is available on our project page. Since the duplication was approximately uniform across entries, the impact on reported metrics is negligible.} via human-in-the-loop validation.
This benchmark is specifically designed for long-form video reasoning with an average video duration of approximately 1,688 seconds.
The duration distribution is concentrated in the 15-30 minute range (71.84\%), with the remaining 28.16\% of videos being longer than 30 minutes.

\section{Training Strategy}
\label{sec:method}

To make full use of the VideoSIAH and elicit robust ``Thinking with Long Videos'' behaviors, LongVT adopts a three-stage training pipeline: \textbf{(1)} cold-start supervised fine-tuning, which teaches the base model to propose temporal windows, invoke video tools, and compose multimodal evidence; \textbf{(2)} agentic reinforcement learning, which optimizes a joint answer–temporal-grounding reward to refine tool-using rollouts; and \textbf{(3)} agentic reinforcement fine-tuning, which distills high-quality RL trajectories back into supervised data to stabilize these behaviors and consolidate long-horizon reasoning.

\subsection{Cold-Start Supervised Fine-Tuning}

\begin{figure*}[t]
  \centering
  \setlength{\abovecaptionskip}{2pt}
  \includegraphics[width=\textwidth]{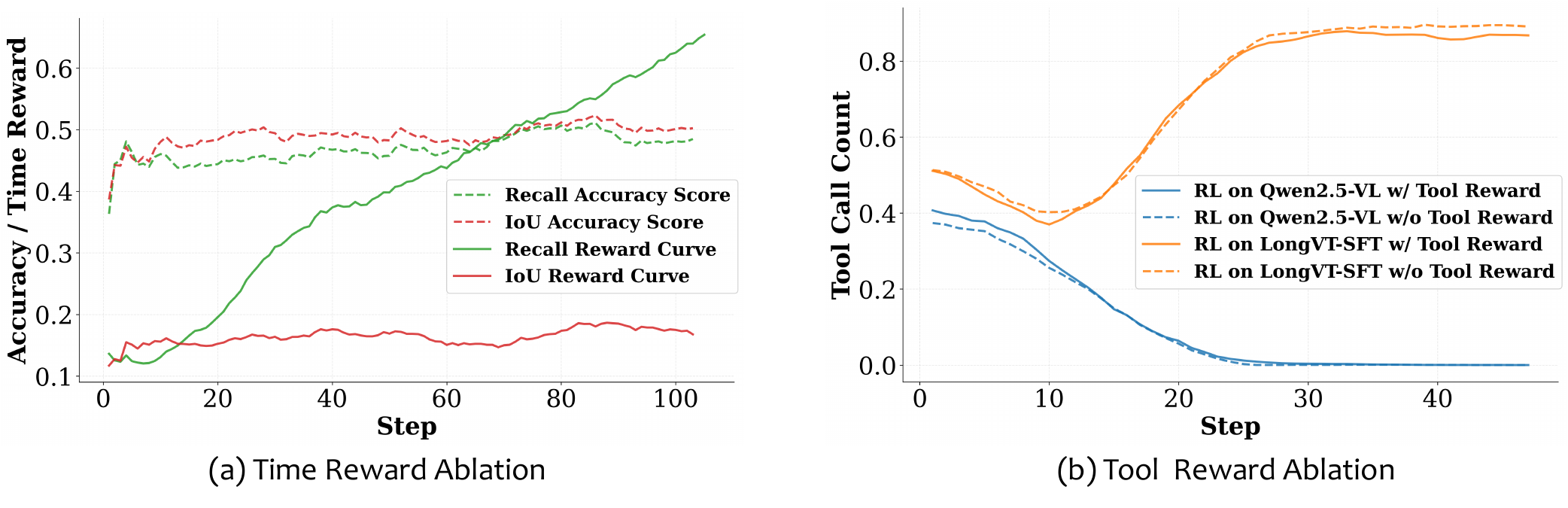}
  \caption{
    \textbf{Ablations on Reward Design.}
    The left panel shows training dynamics under different accuracy and time rewards,
    and the right panel shows the effect of tool-call reward on tool usage.
  }
  \label{fig:reward_grid}
  \setlength{\belowcaptionskip}{-10pt}
\end{figure*}

As shown in \Cref{fig:reward_grid}-(b), our preliminary RL experiments using Qwen2.5-VL-7B \cite{bai2025qwen25vl} as the baseline model reveal that the model fails to improve during RL and ultimately collapses with continued training.
This analysis of training dynamics indicates two major deficiencies of the base LMM: \textbf{(1)} the inability to correctly localize the relevant temporal window within a long video, and \textbf{(2)} insufficient reasoning capability when integrating tool outputs.
We also present a straightforward failure case in the Supplementary Material that illustrates the necessity of a cold-start SFT stage.
These limitations highlight that the model's native tool-calling abilities are too weak for direct RL training. Therefore, a cold-start stage is indispensable for establishing a reliable foundation.
After applying SFT cold start, the model's tool-calling activeness improves substantially and continues to increase steadily during RL, supported by results in \Cref{tab:ablation_exp}.

\subsection{Agentic Reinforcement Learning}
In this stage, we treat the model as a tool-using agent that decides when to inspect the video, how long to crop, and how to integrate the retrieved evidence into its reasoning. 
We employ GRPO \citep{shao2024grpo} to achieve this objective.  
In addition, we introduce a three-part reward modeling that jointly optimizes answer accuracy, format compliance, and temporal grounding precision of sampled trajectories, namely, \emph{joint answer-temporal grounding reward}.
Prior work \cite{feng2025videor1,wang2025timer1} typically targets either answer correctness or time alignment in isolation.
We take a further step toward unifying these signals within a single reward function for open-ended long-video QA.
This coupling ties answer selection to where the evidence lies in time, improving final-answer correctness and promoting more effective tool use at inference, with more reliable and precise timestamp proposals.

\noindent\textbf{Answer Accuracy.}
Let $K$ be the number of sampled rollouts in a group. For the $k$-th rollout ($k\in\{1,\dots,K\}$), let $\hat a^{(k)}$ denote its generated answer and let $a^\star$ denote the ground-truth answer.
Since open-ended QAs cannot be reliably evaluated by rule-based matching, we employ LLM-as-a-Judge \cite{yang2025qwen3} with a strict judging protocol that avoids rewarding ambiguous cases to obtain a categorical verdict
\begin{compactmath}
\[
J^{(k)}=\operatorname{Judge}_{\mathrm{LLM}}\!\big(\hat a^{(k)},\,a^\star\big)\in\{\textsc{F},\textsc{P},\textsc{I}\},
\]
\end{compactmath}
where \textsc{F} = fully consistent (semantically equivalent to $a^\star$), \textsc{P} = partially consistent (contains some correct information but is incomplete or imprecise), and \textsc{I} = inconsistent (incorrect or contradictory).

The accuracy reward is then defined as the normalized score
\begin{compactmath}
\[
\textbf{R}_\text{acc}^{(k)}=
\begin{cases}
1,   & \text{if } J^{(k)}=\textsc{F},\\[2pt]
0.5, & \text{if } J^{(k)}=\textsc{P},\\[2pt]
0,   & \text{if } J^{(k)}=\textsc{I}.
\end{cases}
\]
\end{compactmath}

\noindent\textbf{Format Compliance.}
Let \(y^{(k)}\) denote the full textual output of the \(k\)-th rollout. We set \(\textbf{R}_\text{format}^{(k)}=1\) if \(y^{(k)}\) matches the required output schema \(\mathcal{S}\), and \(0\) otherwise.

\noindent\textbf{Temporal Overlap.}
Following previous temporal grounding work \cite{feng2025videor1,li2025videochatr1}, we use standard temporal IoU as the reward function for temporal localization. For a prediction \([t_s,t_e]\) and ground truth \([t'_s,t'_e]\),
\begin{compactmath}
\[
\mathrm{IoU} \;=\; \frac{\left|[t_s,t_e]\cap[t'_s,t'_e]\right|}{\left|[t_s,t_e]\cup[t'_s,t'_e]\right|}.
\]
\end{compactmath}
\noindent We directly set \(\textbf{R}_\text{time}^{(k)} = \mathrm{IoU}^{(k)}\), which equals \(1\) only when the predicted span matches the ground truth exactly and \(0\) when there is no overlap.

\noindent\textbf{Overall Reward.}
The final reward combines all three components: \(\textbf{R}^{(k)} = \textbf{R}_\text{acc}^{(k)} + \textbf{R}_\text{format}^{(k)} + \textbf{R}_\text{time}^{(k)}\).

\subsection{Agentic Reinforcement Fine-tuning}
We further leverage RFT \cite{sun2025rft} to stabilize the model's agentic behaviors and consolidate multimodal reasoning.
Specifically, we select high-quality cases from early RL rollouts that exhibit both accurate temporal localization and coherent reasoning toward the final answer, and incorporate these trajectories back into the supervised fine-tuning curriculum as privileged and self-distilled demonstrations.
Empirically (see \Cref{sec:ablation}), we find that learning from these in-distribution high-quality trajectories helps the model internalize robust grounding and tool-calling patterns complementary to large-scale agentic RL, effectively guiding optimization toward policies that better align answer accuracy, temporal grounding, and tool usage.

\noindent\textbf{Overall Framework.}
As illustrated in the Supplementary Material, LongVT operates in an iterative ``hypothesis-verification'' cycle: SFT teaches the model to skim global frames and invoke the \texttt{crop\_video} tool to resample fine-grained evidence, and RL consolidates this trajectory via the \emph{joint answer-temporal grounding reward}, enabling learned self-correction when initial retrieval proves insufficient.

\begin{table*}[t]
\centering
\resizebox{\textwidth}{!}{%
\begin{tabular}{l | cc | cccccc | c}
\toprule[1pt]
\multicolumn{1}{c|}{\multirow{2}{*}{\textbf{Model}}} & 
\multicolumn{1}{c}{\textbf{Reasoning}} & 
\multicolumn{1}{c}{\textbf{Tool}} & 
\multicolumn{1}{c}{\textbf{VideoMME ($\approx$1018 sec)} \cite{fu2025videomme}} & 
\multicolumn{3}{c}{\textbf{VideoMMMU ($\approx$506 sec)} \cite{hu2025videommmu}} &
\multicolumn{1}{c}{\textbf{LVBench} \cite{wang2025lvbench}} & 
\multicolumn{1}{c}{\textbf{VideoSIAH-Eval}} & 
\multicolumn{1}{c}{\textbf{Average}} \\
\cmidrule(lr){4-4}\cmidrule(lr){5-7}
& \textbf{Prompt} & \textbf{Calling} & w/ subtitle & adaptation & comprehension & perception & \textbf{($\approx$4101 sec)} & \textbf{($\approx$1688 sec)} & \textbf{Score} \\

\midrule\midrule
\multicolumn{10}{c}{\textbf{\textit{Proprietary LMMs}}} \\
\midrule
GPT-4o \cite{hurst2024gpt4o} & \ding{55} & \ding{55} & \textcolor{gray}{77.2}\textsuperscript{$^\dagger$} & \textcolor{gray}{66.0}\textsuperscript{$^\dagger$} & \textcolor{gray}{62.0}\textsuperscript{$^\dagger$} & \textcolor{gray}{55.7}\textsuperscript{$^\dagger$} & \textcolor{gray}{30.8}\textsuperscript{$^\dagger$} & \textcolor{gray}{17.4} & \textcolor{gray}{51.5} \\ 
Gemini 1.5 Pro \cite{team2024gemini15} & \ding{55} & \ding{55} & \textcolor{gray}{81.3}\textsuperscript{$^\dagger$} & \textcolor{gray}{59.0}\textsuperscript{$^\dagger$} & \textcolor{gray}{53.3}\textsuperscript{$^\dagger$} & \textcolor{gray}{49.3}\textsuperscript{$^\dagger$} & \textcolor{gray}{33.1}\textsuperscript{$^\dagger$} & - & \textcolor{gray}{55.2} \\ 

\midrule
\multicolumn{10}{c}{\textbf{\textit{Open-Source LMMs with Sparse Frame Sampling}}} \\
\midrule
Qwen2.5-VL-7B \cite{bai2025qwen25vl} & \ding{55} & \ding{55} & \underline{62.6} & \underline{37.3} & 28.0 & 36.7 & 30.7 & \underline{28.1} & 37.2 \\ 
Video-R1-7B \cite{feng2025videor1} & \ding{51} & \ding{55} & 61.0 & 36.3 & 40.7 & 52.3 & 37.2 & 27.9 & \underline{42.6} \\
VideoRFT-7B \cite{wang2025videorft} & \ding{51} & \ding{55} & 60.9 & 36.7 & 42.0 & \underline{53.0} & 34.7 & 26.5 & 42.3 \\ 
Video-Thinker-7B \cite{wang2025videothinker} & \ding{51} & \ding{55} & 61.0 & 34.3 & \underline{44.7} & \underline{53.0} & \textbf{52.2} & 10.4 & \underline{42.6} \\ 
\rowcolor{gray!20} LongVT-7B-SFT (Ours) & \ding{51} & \ding{51} & 12.5 & \textbf{37.7} & \textbf{46.0} & \textbf{58.3} & 36.0 & 26.8 & 36.2 \\ 
\rowcolor{gray!20} \textbf{LongVT-7B-RL (Ours)} & \ding{51} & \ding{51} & \textbf{66.1} & 32.7 & \underline{44.7} & 50.0 & \underline{37.8} & \textbf{31.0} & \textbf{43.7} \\ 

\midrule
\multicolumn{10}{c}{\textbf{\textit{Open-Source LMMs with Dense Frame Sampling}}} \\
\midrule
Qwen2.5-VL-7B \cite{bai2025qwen25vl} & \ding{55} & \ding{55} & 64.3 & 35.7 & \textbf{44.3} & 54.7 & 40.9 & 33.8 & 46.0 \\ 
Video-R1-7B \cite{feng2025videor1} & \ding{51} & \ding{55} & 60.5 & \underline{37.3} & 38.7 & 46.3 & 40.1 & 33.1 & 42.7 \\ 
VideoRFT-7B \cite{wang2025videorft} & \ding{51} & \ding{55} & 49.2 & \textbf{37.7} & 40.7 & 48.7 & 18.7 & 26.9 & 37.0 \\ 
Video-Thinker-7B \cite{wang2025videothinker} & \ding{51} & \ding{55} & 60.8 & \textbf{37.7} & 42.7 & 55.3 & \textbf{54.3} & 6.6 & 42.9 \\ 
\rowcolor{gray!20} LongVT-7B-SFT (Ours) & \ding{51} & \ding{51} & 64.9 & 32.3 & 42.0 & 49.7 & 41.1 & 34.8 & 44.1 \\ 
\rowcolor{gray!20} LongVT-7B-RL (Ours) & \ding{51} & \ding{51} & \underline{66.1} & \textbf{37.7} & 42.3 & \underline{56.3} & \underline{41.4} & \underline{35.9} & \underline{46.6} \\ 
\rowcolor{gray!20} \textbf{LongVT-7B-RFT (Ours)} & \ding{51} & \ding{51} & \textbf{67.0} & 35.7 & \underline{43.7} & \textbf{56.7} & 41.3 & \textbf{42.0} & \textbf{47.7} \\ 

\bottomrule[1pt]
\end{tabular}
}
\caption{\textbf{Performance Comparison with Existing Video-Centric LMMs across Various Long Video Understanding and Reasoning Benchmarks.} 
The best and second-best result among open-source models in each column is marked in \textbf{bold} and \underline{underlined}, respectively.
The numbers with ``$\approx$'' denote the average video duration of each benchmark.
$^\dagger$ indicates results sourced from official reports \cite{fu2025videomme,hu2025videommmu,wang2025lvbench}.}
\label{tab:main_exp}
\end{table*}

\section{Experiments}
\label{sec:exp}

\subsection{Experimental Setup}

We utilize Qwen2.5-VL-7B \cite{bai2025qwen25vl} as the baseline model in all experiments.
We report the performance of three LongVT variants based on their training stages against Qwen2.5-VL-7B and other open-source video-centric LMMs including Video-R1-7B \cite{feng2025videor1}, VideoRFT-7B \cite{wang2025videorft}, and Video-Thinker-7B \cite{wang2025videothinker} plus proprietary LMMs such as GPT-4o \cite{hurst2024gpt4o} and Gemini 1.5 Pro \cite{team2024gemini15}.
Note that we do not include direct comparisons to the concurrent tool-augmented video-centric LMM \cite{zhang2025vital}, since its model checkpoints are not publicly available, which hinders fair and reproducible experiments.
We evaluate all models on four long-video understanding and reasoning benchmarks, namely VideoMME \cite{fu2025videomme}, VideoMMMU \cite{hu2025videommmu}, LVBench \cite{wang2025lvbench}, and our self-curated VideoSIAH-Eval, leveraging a unified evaluation framework \cite{zhang2025lmmseval} for fair comparison.
Results are reported under two frame-sampling regimes: \textit{\textbf{Sparse Frame Sampling}} (64 uniformly sampled video frames) and \textit{\textbf{Dense Frame Sampling}} (512 or 768 uniformly sampled frames; the better result of the two is reported).
\textbf{Reasoning Prompt} indicates whether a standard reasoning-style prompt (\ding{51}) or a direct question-answering prompt (\ding{55}) is applied; \textbf{Tool Calling} denotes whether native tool calling is enabled (\ding{51}) or disabled (\ding{55}) in the prompt. More implementation details can be found in the Supplementary Material.

\subsection{Main Results}

As shown in \Cref{tab:main_exp}, our approach achieves a new state-of-the-art among open-source video-centric LMMs under both sparse and dense frame sampling settings.
When evaluating at 64 frames, LongVT-7B-RL slightly surpasses the best existing open-source baseline.
Under dense frame sampling, both LongVT-7B-RL and LongVT-7B-RFT yield more dominant performance, outperforming existing methods by a large margin.
On the challenging VideoSIAH-Eval, which involves open-ended QAs that require the retrieval of fine-grained evidence from hours-long videos, LongVT-7B-RFT reaches 42.0, outperforming the second-best model by 6 points.
This confirms that LongVT achieves stronger long-video reasoning and exhibits an emergent ability to invoke native tools for temporal localization.
Notably, the gap between open-source and proprietary LMMs has narrowed substantially: LongVT's best-performing checkpoint lies within roughly four points of GPT-4o on average, marking a significant step forward in long-video reasoning capability among open-source LMMs.
Despite incorporating multi-turn tool interactions, LongVT incurs no additional inference latency and can even be faster than single-turn baselines by avoiding hallucination-driven verbose generation; we provide a detailed efficiency analysis in the Supplementary Material.
\begin{table*}[t]
\centering
\resizebox{\textwidth}{!}{%
\begin{tabular}{l | cccccc | c}
\toprule[1pt]
\multicolumn{1}{c|}{\multirow{2}{*}{\textbf{Setting}}} &
\multicolumn{1}{c}{\textbf{VideoMME} \cite{fu2025videomme}} &
\multicolumn{3}{c}{\textbf{VideoMMMU} \cite{hu2025videommmu}} &
\multicolumn{1}{c}{\textbf{LVBench} \cite{wang2025lvbench}} &
\multicolumn{1}{c}{\textbf{VideoSIAH-Eval}} &
\multicolumn{1}{c}{\textbf{Average}}\\
\cmidrule(lr){2-2}\cmidrule(lr){3-5}\cmidrule(lr){6-6}\cmidrule(lr){7-7}
& w/ subtitle & adaptation & comprehension & perception & test & test & \textbf{Score} \\
\midrule\midrule
\multicolumn{8}{c}{\textbf{\textit{Data Recipe}}} \\
\midrule
SFT w/o self-curated iMCoTT & 8.4 & \textbf{33.6} & 41.6 & 46.0 & 15.1 & 4.1 & 24.8 \\
\rowcolor{gray!20} SFT w/ self-curated iMCoTT (LongVT-7B-SFT) & \textbf{64.9} & 32.3 & \textbf{42.0} & \textbf{49.7} & \textbf{41.1} & \textbf{34.8} & \textbf{44.1} \\
RL w/o self-curated QAs  & 55.1 & 30.6 & 42.0 & 45.6 & 38.4 & 30.8 & 40.4 \\
\rowcolor{gray!20} RL w/ self-curated QAs (LongVT-7B-RL) & \textbf{66.1} & \textbf{37.7} & \textbf{42.3} & \textbf{56.3} & \textbf{41.4} & \textbf{35.9} & \textbf{46.6} \\
\midrule
\multicolumn{8}{c}{\textbf{\textit{Training Stage}}} \\
\midrule
SFT only (LongVT-7B-SFT) & 64.9 & 32.3 & 42.0 & 49.7 & 41.1 & 34.8 & 44.1 \\
RL only & 52.7 & 35.3 & 43.0 & 55.1 & 37.1 & 28.2 & 41.9 \\
SFT+RL (LongVT-7B-RL) & 66.1 & \textbf{37.7} & 42.3 & 56.3 & \textbf{41.4} & 35.9 & 46.6 \\
\rowcolor{gray!20} SFT+RL+RFT (LongVT-7B-RFT) & \textbf{67.0} & 35.7 & \textbf{43.7} & \textbf{56.7} & 41.3 & \textbf{42.0} & \textbf{47.7} \\
\midrule
\multicolumn{8}{c}{\textbf{\textit{Decoupled Temporal Grounding Reward}}} \\
\midrule
\multirow{2}{*}{}
  & 
  & \multicolumn{4}{c}{\textbf{Charades-STA} \cite{gao2017charades}}
  & 
  & \textbf{Average} \\
\cmidrule(lr){3-6}
  &  
  & \multicolumn{1}{c}{IoU@0.3}
  & \multicolumn{1}{c}{IoU@0.5}
  & \multicolumn{1}{c}{IoU@0.7}
  & \multicolumn{1}{c}{mIoU}
  & 
  & \textbf{Score} \\
\midrule
RL w/o Decoupled Reward & & 31.5 & 19.9 & 9.1 & 21.2 & & 20.4 \\
RL w/ Recall Reward     & & 32.0 & 20.4 & 9.6 & 21.6 & & 20.9 \\
\rowcolor{gray!20} RL w/ IoU Reward & & \textbf{41.0} & \textbf{25.8} & \textbf{11.7} & \textbf{27.2} & & \textbf{26.4} \\
\bottomrule[1pt]
\end{tabular}
}
\caption{\textbf{Ablation Studies.} 
The best result among each comparison group is in \textbf{bold}.
We examine \textbf{\textit{Data Recipe}} where we remove self-curated iMCoTTs during SFT or self-curated QAs during RL to test the dependence on fine-grained supervision; \textbf{\textit{Training Stage}} where SFT, RL, and RFT are ablated individually and in combination to test their complementary effect; \textbf{\textit{Decoupled Temporal Grounding Reward}} where Recall-based and IoU-based reward functions are compared, together with a variant without decoupled temporal grounding reward.}
\label{tab:ablation_exp}
\end{table*}

\subsection{Ablation Studies}
\label{sec:ablation}

\noindent\textbf{Fine-grained reasoning data matters.} As shown in \Cref{tab:ablation_exp}, our self-curated training data plays a crucial role in shaping the model's reasoning behavior when dealing with long-form videos.
In the SFT stage, removing the self-curated iMCoTTs (SFT w/o self-curated iMCoTT) leads to a consistent performance drop in long-form video understanding.
In addition, when self-curated QAs are removed during RL (RL w/o self-curated QAs), the model's performance drops quickly on VideoSIAH-Eval, with lower answer accuracy, weaker temporal localization, and less systematic tool use, which can also be observed in \Cref{fig:reward_grid}-(b).

\noindent\textbf{Recall encourages coverage; IoU demands precision.} As shown in \Cref{fig:reward_grid}-(a), using Recall as the reward function during RL presents a drawback: the policy can enlarge the predicted span to envelop the ground-truth interval, which monotonically raises the Recall-based score while ignoring boundary quality.
This plateau in the curve of Recall Accuracy Score further validates our hypothesized reward hacking.
Quantitatively, in the reward-choice rows of \Cref{tab:ablation_exp}, IoU-rewarded training outperforms Recall-rewarded training on the temporal grounding benchmark \cite{gao2017charades}, while Recall is only marginally above the RL w/o Decoupled Reward variant, pointing to IoU's tighter handling of boundary agreement.
Optimizing with IoU provides smooth shaping over overlap and implicitly penalizes span inflation via the union term, yielding better-aligned boundaries and more disciplined tool use.

\noindent\textbf{Is tool reward really necessary?} As shown in \Cref{fig:reward_grid}-(b), the Qwen2.5-VL-7B baseline collapses to near-zero tool calls after training in both configurations (w/ and w/o tool reward), indicating that the model does not internalize the tool's function.
After performing cold-start SFT to obtain LongVT-7B-SFT, tool-call frequency rises during training under both configurations and accuracy improves in tandem.
Hence, the tool reward is not required for basic competence: once SFT grounds the tool's semantics, the model learns when to invoke the tool and when to abstain.
Moreover, introducing the tool reward brings little benefit.
In the later training stage, the configuration without the tool reward even exhibits slightly higher tool-use frequency, indicating that the binary bonus does not encourage usage and may suppress exploration, while accuracy remains essentially unchanged.
Given these observations, we discard the tool reward in our final recipe and rely on the standard accuracy, format, and decoupled IoU reward modeling.

\noindent\textbf{SFT builds competence; RL optimizes decisions; RFT stabilizes behaviors.}  
We ablate each training stage individually and in combination, finding that strong performance emerges only with the full three-stage pipeline. As shown in \Cref{fig:reward_grid}-(b), removing SFT leaves the model with poor tool-use ability: it cannot reliably invoke the \texttt{crop\_video} tool or integrate cropped evidence into its reasoning. Consistently, the RL-only variant achieves the lowest scores on all four benchmarks (\Cref{tab:ablation_exp}) and exhibits behavioral inconsistencies during training\textemdash{}often following surface instructions and becoming confused by the returned crop rather than using it as supporting evidence.

SFT teaches the intended tool-use paradigm—selecting temporal windows, inspecting their content, and incorporating the resulting evidence into the final answer.
However, SFT remains imitation-driven \cite{li2024getting}: it fits demonstrated formats, suffers from exposure bias, and fails to generalize under distribution shift.
On long-video QA, SFT alone yields only modest gains. We therefore introduce RL with a temporal-grounding reward, optimized via GRPO.
RL enables the policy to learn \emph{when} to inspect, \emph{how long} to crop, and \emph{how} to integrate retrieved evidence. This stage pushes performance beyond the supervised ceiling on held-out videos and unseen question templates (\Cref{tab:ablation_exp}), aligning with prior findings that GRPO improves reasoning and generalization \cite{guo2025deepseekr1}.

Finally, RFT distills high-reward trajectories back into the supervised corpus, providing additional performance gains. On VideoSIAH-Eval, it surpasses the RL-only plateau by a substantial margin and yields our best-performing model, while still delivering consistent improvements on other benchmarks. This demonstrates that consolidating successful rollouts is essential for fully realizing the benefits of temporal-grounding feedback.

\section{Conclusion}
\label{sec:conclusion}

In this work, we present \textbf{LongVT}, an end-to-end agentic framework that enables LMMs to reliably reason over long videos.
By interleaving multimodal tool-augmented CoT with on-demand temporal inspection, LongVT transforms long-video understanding from passive frame consumption into active, evidence-seeking reasoning.
Supported by self-curated \textbf{VideoSIAH}, a large-scale, fine-grained data suite built specifically for evidence-sparse long-video reasoning tasks, our proposed three-stage training pipeline yields substantial and consistent improvements compared to existing strong baselines.

\section*{Acknowledgements}
This project was fully supported by MiroMind, which provided the compute, storage, and engineering infrastructure used for all experiments reported in this paper. 

{
    \small
    \bibliographystyle{ieeenat_fullname}
    \bibliography{main}
}

\clearpage
\setcounter{page}{1}
\setcounter{section}{0}
\setcounter{table}{0}
\setcounter{figure}{0}
\setcounter{footnote}{0}
\maketitlesupplementary

\centerline{\large{\textbf{Outline}}}
This Supplementary Material complements the main paper, providing comprehensive experimental details, in-depth analyses of training dynamics, and extensive qualitative visualizations. The content is organized as follows:

\begin{itemize}[leftmargin=*]
    \item \textbf{Strategic Alignment \& Motivation.} We first demonstrate the conceptual alignment between LongVT and state-of-the-art proprietary large multimodal models (LMMs) in \Cref{sec:sota_case}. Subsequently, we present a rigorous data contamination study in \Cref{sec:motivation_videosiah} to underscore the necessity of our proposed VideoSIAH-Eval benchmark, followed by detailed statistics of the curated dataset in \Cref{sec:detail_videosiah}.
    
    \item \textbf{Formulation \& Training Dynamics.} We present the overall framework illustration in \Cref{fig:framework} and elaborate on the theoretical formulations of our training objectives in \Cref{sec:detail_method} for both supervised fine-tuning (SFT) and reinforcement learning (RL). Crucially, in \Cref{sec:reflection_stat}, we visualize the ``economy of thinking''\textemdash{}a distinct evolutionary trajectory where the model learns to internalize tool usage. \Cref{sec:detail_impl} then provides the exact hyperparameters and infrastructure details for reproducibility.
    
    \item \textbf{Efficiency \& Qualitative Analysis.} We report a detailed inference latency comparison in \Cref{sec:efficiency}, countering the intuition that multi-turn agentic frameworks are inherently slower. In \Cref{sec:examples}, we provide prompt templates, diverse qualitative examples, and workflow demonstration, while \Cref{sec:failure_case} analyzes specific failure modes to highlight the importance of the cold-start training stage.
    
    \item \textbf{Discussion.} Finally, we discuss the architectural limitations and future multi-agent directions in \Cref{sec:limitation}, followed by a discussion on the broader impact and ethical considerations in \Cref{sec:impact} and \Cref{sec:ethics}, respectively.
\end{itemize}

\section{LongVT Performs Human-Aligned Thinking like Leading Proprietary LMMs}
\label{sec:sota_case}

\begin{figure*}[t]
\centering
\begin{subfigure}{0.405\linewidth}
\centering
\includegraphics[width=\linewidth]{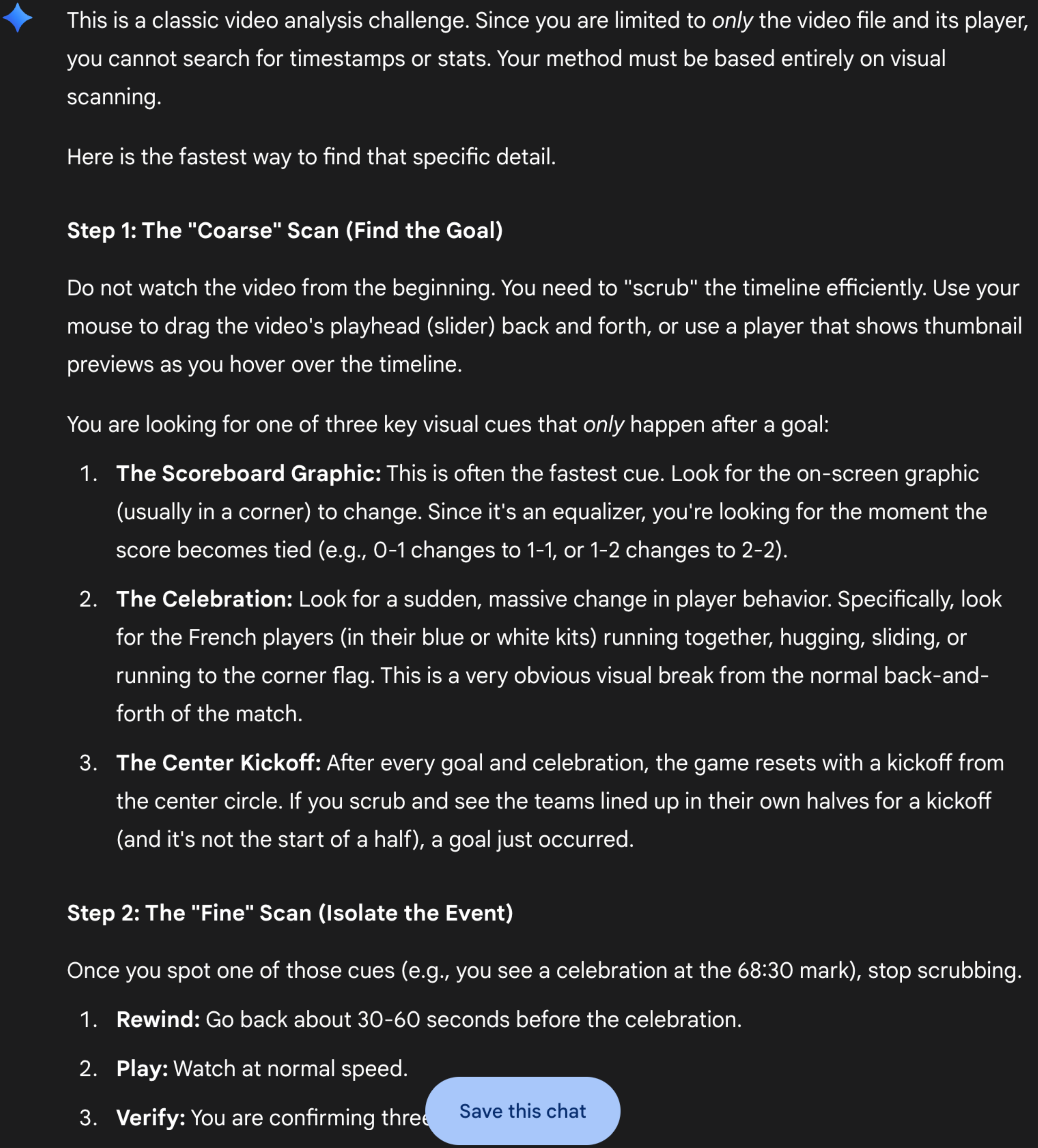}
\caption{Watching Strategy of Gemini 2.5 Pro.}
\label{fig:gemini_2_5}
\end{subfigure}
\hfill
\begin{subfigure}{0.555\linewidth}
\centering
\includegraphics[width=\linewidth]{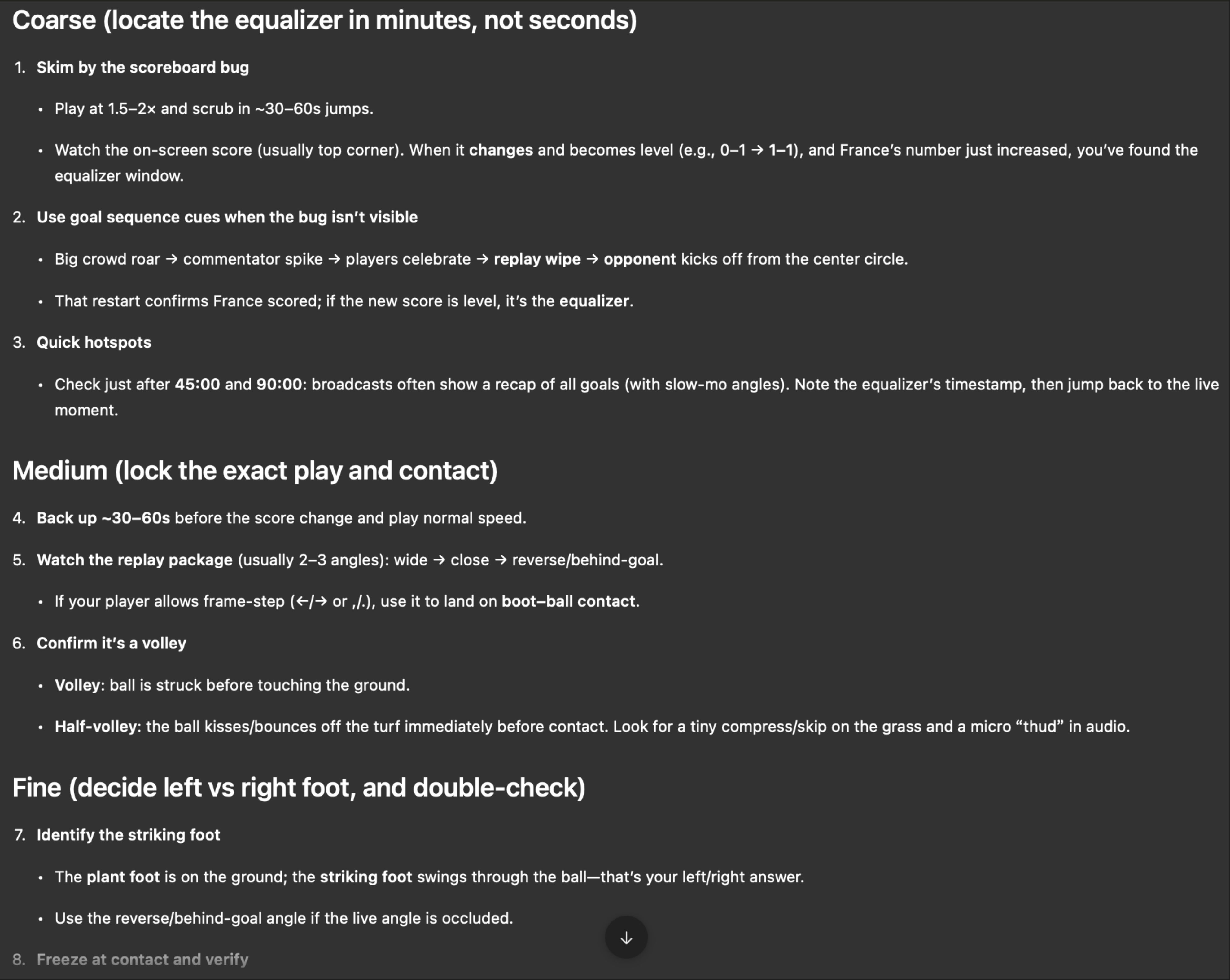}
\caption{Watching Strategy of GPT-5 Thinking.}
\label{fig:gpt_5}
\end{subfigure}
\caption{\textbf{Comparison of Watching Strategies Proposed by Gemini 2.5 Pro \cite{comanici2025gemini25} and GPT-5 Thinking \cite{openai2025gpt5}.} Best viewed when zoomed in.}
\label{fig:watching_strategies}
\end{figure*}

The core philosophy of our proposed interleaved Multimodal Chain-of-Tool-Thought (iMCoTT) entails a ``global-to-local'' thinking pattern: the model first performs a coarse skim to formulate a hypothesis, and subsequently invokes the native \texttt{crop\_video()} tool to inspect specific temporal windows for fine-grained verification.
While this design was inspired by human intuition, we observe a striking convergence between our approach and the reasoning behaviors emerging in state-of-the-art proprietary LMMs when they are prompted to perform fine-grained analysis.

To validate this alignment, we queried two leading models, Gemini 2.5 Pro~\cite{comanici2025gemini25} and GPT-5 Thinking~\cite{openai2025gpt5}, regarding their optimal strategies for analyzing fine-grained video details. As illustrated in \Cref{fig:gemini_2_5}, Gemini 2.5 Pro explicitly advocates for a two-stage process: a ``Step 1: Coarse Scan'' to efficiently locate the general event (e.g., searching for scoreboard changes or crowd reactions), followed by a ``Step 2: Fine Scan'' to isolate the exact moment and verify details (e.g., scrubbing back 30-60 seconds). 
This directly mirrors the workflow of our proposed LongVT, where the ``Coarse Scan'' corresponds to our global preview stage, and the ``Fine Scan'' is functionally identical to our agentic \texttt{crop\_video()} tool calling.
Similarly, \Cref{fig:gpt_5} demonstrates that the GPT-series model adopts a hierarchical ``Coarse$\rightarrow$Medium$\rightarrow$Fine'' search strategy. 
These examples confirm that the ``Thinking with Long Videos'' paradigm we propose in this work is a natural and necessary evolution for reliable long-form video reasoning, given that such human-aligned reasoning capabilities are currently exclusive to top-tier proprietary models.

\section{What Motivates VideoSIAH? Unveiling the Data Contamination in Qwen-VL Series}
\label{sec:motivation_videosiah}

\begin{table*}[t]
\centering
\small
\setlength{\tabcolsep}{6pt}
\begin{tabular}{l c ccc c}
\toprule[1pt]
\multirow{2}{*}{\textbf{Setting}} & \textbf{VideoMME}~\cite{fu2025videomme} & \multicolumn{3}{c}{\textbf{VideoMMMU}~\cite{hu2025videommmu}} & \textbf{VideoSIAH-Eval} \\
\cmidrule(lr){2-2} \cmidrule(lr){3-5} \cmidrule(lr){6-6}
& w/o subtitle & adaptation\textsuperscript{$\ast$} & comprehension & perception & test \\
\midrule
\multicolumn{6}{c}{\textbf{\textit{Qwen2.5-VL-7B-Instruct}} \cite{bai2025qwen25vl}} \\
\midrule
\rowcolor{gray!15} 
Original & \textbf{64.3} & \textbf{35.7} & \textbf{44.3} & \underline{54.7} & \textbf{33.8} \\
No Visual & 40.1 & 27.0 & 38.3 & 39.3 & \underline{12.7} \\
Rearranged Choices & \underline{56.0} & \underline{31.6} & \underline{40.3} & \textbf{67.0} & - \\
\midrule
\multicolumn{6}{c}{\textbf{\textit{Qwen3-VL-8B-Instruct}} \cite{qwen2025qwen3vl}} \\
\midrule
\rowcolor{gray!15} 
Original & \textbf{69.3} & \textbf{40.7} & \textbf{60.3} & \textbf{71.3} & \textbf{46.6} \\
No Visual & 44.1 & 35.1 & 39.3 & 46.7 & \underline{0.00} \\
Rearranged Choices & \underline{69.0} & \underline{38.7} & \underline{47.7} & \underline{69.3} & - \\
\bottomrule[1pt]
\end{tabular}
\caption{\textbf{Contamination Tests for Qwen-VL Series on Long Video Understanding and Reasoning Benchmarks.} Results are reported across different perturbation settings. The best result in each block column is in \textbf{bold}, and the second-best is \underline{underlined}. The VideoSIAH-Eval column shows ``-'' entries for Rearranged Choices since our proposed benchmark is fully open-ended QA, where random option-answer mapping is not applicable. Note that adaptation\textsuperscript{$\ast$} is evaluated exclusively on multiple-choice questions.}
\label{tab:contamination}
\end{table*}

With the rapid advancements of LMMs, model performance on various benchmarks has steadily improved. However, the ``black-box'' nature of training data raises a critical question: \emph{Do these improvements reflect genuine reasoning capability, or are they partly due to the model memorizing the benchmark samples?} To investigate this, we conduct a rigorous contamination study on the Qwen-VL series \cite{bai2025qwen25vl,qwen2025qwen3vl} across two probing settings: \textbf{(1)} No Visual, where we feed the text prompt without video frames to test for direct memorization; \textbf{(2)} Rearranged Choices, where we randomize the mapping between option labels and their textual content (e.g., assigning the original answer A to B) for multiple-choice questions (MCQs) to detect label memorization.

Our experimental results reveal significant vulnerabilities in existing benchmarks and highlight the necessity of our proposed VideoSIAH-Eval:
\emph{Observation 1: ``No Visual'' Performance Indicates Severe Leakage in Existing Benchmarks.}
As shown in \Cref{tab:contamination}, both Qwen2.5-VL and Qwen3-VL achieve remarkably high scores on VideoMME and VideoMMMU even without seeing any video frames.
Notably, for VideoMME, we specifically evaluate without subtitles to ensure there is no textual leakage, yet Qwen2.5-VL still achieves 40.1\%, far exceeding random guessing ($\sim$25\%) for such four-option MCQs.
Similar patterns of potential data leakage are observed on VideoMMMU. While the `No Visual' scores of 38.3\% (Comprehension) and 39.3\% (Perception) might appear similar to VideoMME, they are statistically more improbable given the dataset composition. Our statistics reveal that these subsets are overwhelmingly dominated by MCQs with 10 options (e.g., 286 out of 300 for Comprehension and 279 out of 300 for Perception), implying a random guessing baseline of only $\sim10\text{--}16\%$. The fact that the model achieves scores significantly above this threshold absent any visual context indicates a high probability of benchmark memorization.
In contrast, performance on VideoSIAH-Eval drops significantly in the ``No Visual'' setting. Specifically, Qwen3-VL collapses to a score of 0.00. Upon manual inspection, we find that without visual grounding, the model generates repetitive code or refusal messages, which is the expected behavior for a clean and non-contaminated benchmark.
\emph{Observation 2: ``Rearranged Choices'' Reveals Overfitting to Option Patterns.}
For MCQ-based benchmarks, we observe distinct performance drops when answer choices are rearranged. For instance, Qwen2.5-VL drops from 64.3 to 56.0 on VideoMME. This indicates that they heavily rely on memorizing specific option mappings (e.g., the answer to this question is usually ``A'') rather than understanding the content. Since VideoSIAH-Eval utilizes a fully open-ended QA format, it is inherently immune to this type of option hacking, providing a more robust assessment of the model's capabilities.

These findings confirm that existing benchmarks are compromised by data contamination (high ``No Visual'' scores) and option bias (sensitive to ``Rearranged Choices''). This motivates the introduction of VideoSIAH-Eval, which ensures: (1) \emph{Zero leakage} as verified by the 0.00 blind score, and (2) \emph{Immunity to option bias} via open-ended QA format.

\section{Additional VideoSIAH Details}
\label{sec:detail_videosiah}

\begin{figure*}[t]
\centering
\begin{subfigure}{0.49\linewidth}
\centering
\includegraphics[width=\linewidth]{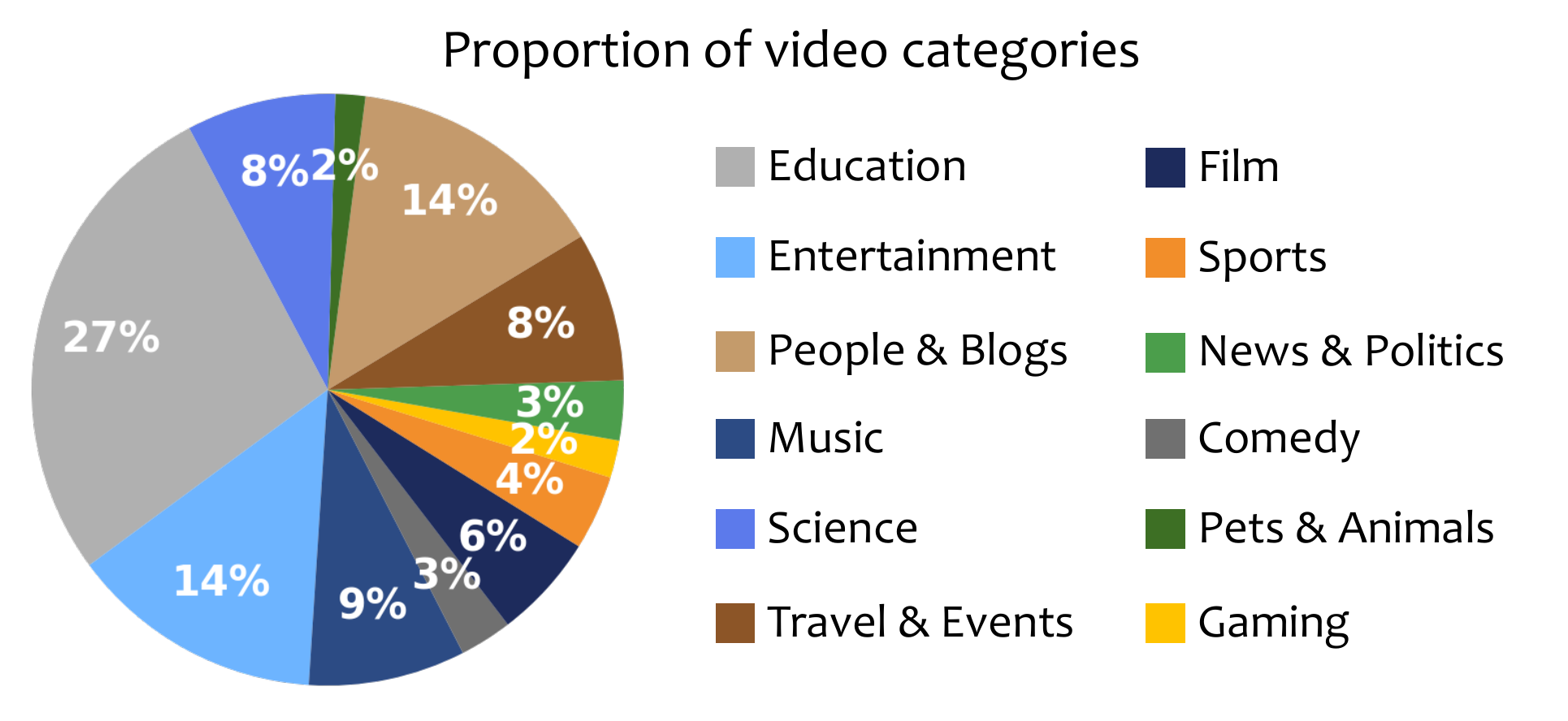}
\caption{Video Category Distribution}
\label{fig:video_category}
\end{subfigure}
\hfill
\begin{subfigure}{0.49\linewidth}
\centering
\includegraphics[width=\linewidth]{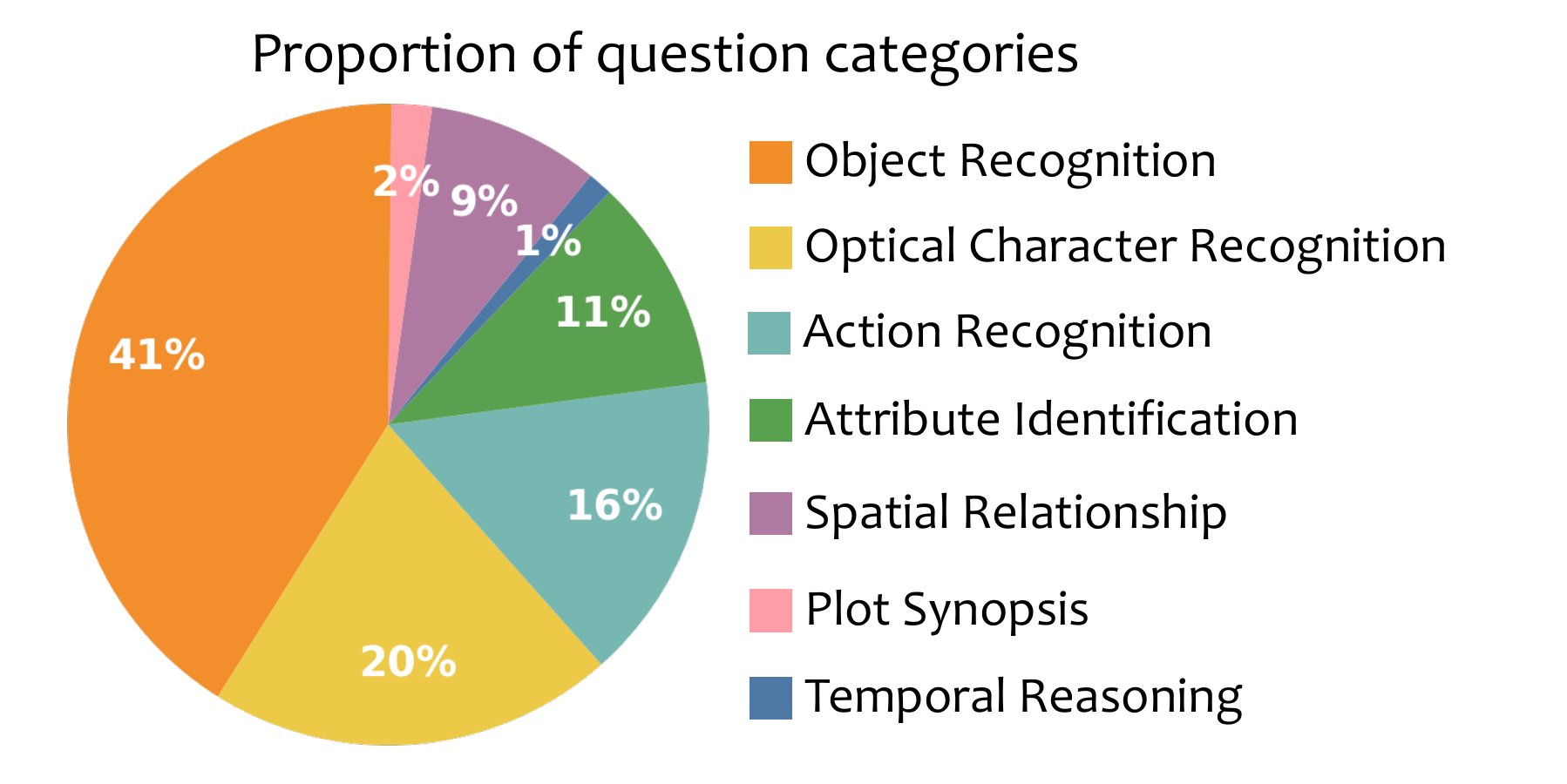}
\caption{Question Category Distribution}
\label{fig:question_category}
\end{subfigure}
\caption{\textbf{Category Distribution of VideoSIAH-Eval.} We present the distribution of video types (a) and question types (b), highlighting the diversity of our proposed benchmark.}
\label{fig:category_distribution}
\end{figure*}

\paragraph{Breakdown of Image-based CoT Data.}

\begin{table}[ht]
\centering
\resizebox{0.45\textwidth}{!}{%
\begin{tabular}{l l r}
\toprule[1pt]
\textbf{Source} & \textbf{Purpose} & \textbf{Samples} \\
\midrule\midrule
LLaVA-CoT \cite{xu2025llavacot} & General Visual Reasoning & 54,591 \\
OpenVLThinker \cite{deng2025openvlthinker} & Complex Reasoning & 2,829 \\
We-Math 2.0 \cite{qiao2025wemath2} & Mathematical Reasoning & 602 \\
\bottomrule[1pt]
\end{tabular}
}
\caption{\textbf{Detailed Statistics of Image-based CoT Data for Cold-Start SFT.}}
\label{tab:image_cot}
\end{table}

As detailed in \Cref{tab:image_cot}, we construct a diverse mixture of image-based CoT data for the cold-start SFT stage, spanning general visual reasoning \cite{xu2025llavacot}, complex logical inference \cite{deng2025openvlthinker}, and mathematical problem-solving \cite{qiao2025wemath2}.
Drawing on insights from recent work \cite{zhang2025videollama3,feng2025videor1}, we leverage these image-based reasoning traces to strengthen the model's fundamental perceptual capabilities. This strategy exploits the inherent synergy between image and video modalities, where robust spatial grounding serves as a critical foundation for complex temporal reasoning.

\paragraph{Category Distribution for VideoSIAH-Eval.}
VideoSIAH-Eval comprises 244 videos and 652 high-quality QA pairs.
As illustrated in \Cref{fig:video_category}, the video corpus encompasses a diverse spectrum of domains, ranging from Travel \& Events to Gaming, ensuring broad coverage of real-world scenarios.
Furthermore, \Cref{fig:question_category} highlights our deliberate emphasis on dynamic video reasoning: Action Recognition and Temporal Reasoning (17\% in total) constitute a large portion of queries, rigorously benchmarking the model's capacity for fine-grained event perception and causal understanding in the temporal dimension.

\section{Additional Methodological Details}
\label{sec:detail_method}

\paragraph{Next-Token Prediction.}
During SFT, we train our model by minimizing the negative log-likelihood of the target tokens given their preceding context.
For a sequence of tokens \(x = (x_{1}, x_{2}, \ldots, x_{T})\) and a model parameterized by \(\theta\) that defines conditional probabilities \(p_{\theta}(x_{t} \mid x_{<t})\), the loss function is defined as
\[
\mathcal{L}(\theta) = - \sum_{t=1}^{T} \log p_{\theta}(x_{t} \mid x_{<t}),
\]
which encourages the model to assign higher probability to the ground-truth next token.

\paragraph{Group Relative Policy Optimization.}
During RL, we adopt GRPO \cite{shao2024grpo} for optimization.
For each prompt \(x\in\mathcal{D}\), we draw a group of \(K\) responses from the behavior policy \(\pi_{\theta_{\mathrm{old}}}\).
\begin{compactmath}
\[
\begin{aligned}
& y^{(k)} \sim \pi_{\theta_{\mathrm{old}}}(\cdot\mid x),\quad k=1,\ldots,K,\\
& y^{(k)}=(y^{(k)}_1,\ldots,y^{(k)}_{T_k}),\qquad T_k=\text{len}(y^{(k)}).
\end{aligned}
\]
\end{compactmath}

We use a group baseline and advantages:
\begin{compactmath}
\[
\begin{aligned}
& b=\frac{1}{K}\sum_{k=1}^K R^{(k)},\qquad
A^{(k)}=R^{(k)}-b,
\end{aligned}
\]
\end{compactmath}
where \(R^{(k)}\) is the scalar return of response \(y^{(k)}\).

The policy maximizes a length-normalized, token-conditional KL-regularized objective:
\begin{equation}\label{eq:grpo_obj}
\resizebox{\columnwidth}{!}{$
\begin{aligned}
\mathcal{J}(\theta)
&\!\!=\mathbb{E}_{\substack{x\sim\mathcal{D}\\ \{y^{(k)}\}\sim \pi_{\theta_{\mathrm{old}}}(\cdot\mid x)}}\!\bigg[
\frac{1}{K}\sum_{k=1}^K \frac{1}{T_k}\sum_{t=1}^{T_k}
A^{(k)} \log \pi_\theta\!\big(y^{(k)}_t \mid x, y^{(k)}_{<t}\big)
\bigg] \\[-1pt]
&\!\!-\beta\,\mathbb{E}_{x\sim\mathcal{D}}\!\bigg[
\frac{1}{K}\sum_{k=1}^K \frac{1}{T_k}\sum_{t=1}^{T_k}
D_{\mathrm{KL}}\!\Big(\pi_\theta(\cdot\!\mid\! x, y^{(k)}_{<t}) \,\Vert\, \pi_{\mathrm{ref}}(\cdot\!\mid\! x, y^{(k)}_{<t})\Big)
\bigg],
\end{aligned}
$}
\end{equation}
with \(t\in\{1,\ldots,T_k\}\), \(\pi_{\mathrm{ref}}\) a frozen reference policy, and \(\beta>0\) controlling KL strength.

\begin{figure*}[t]
    \centering
    \includegraphics[width=0.9\linewidth]{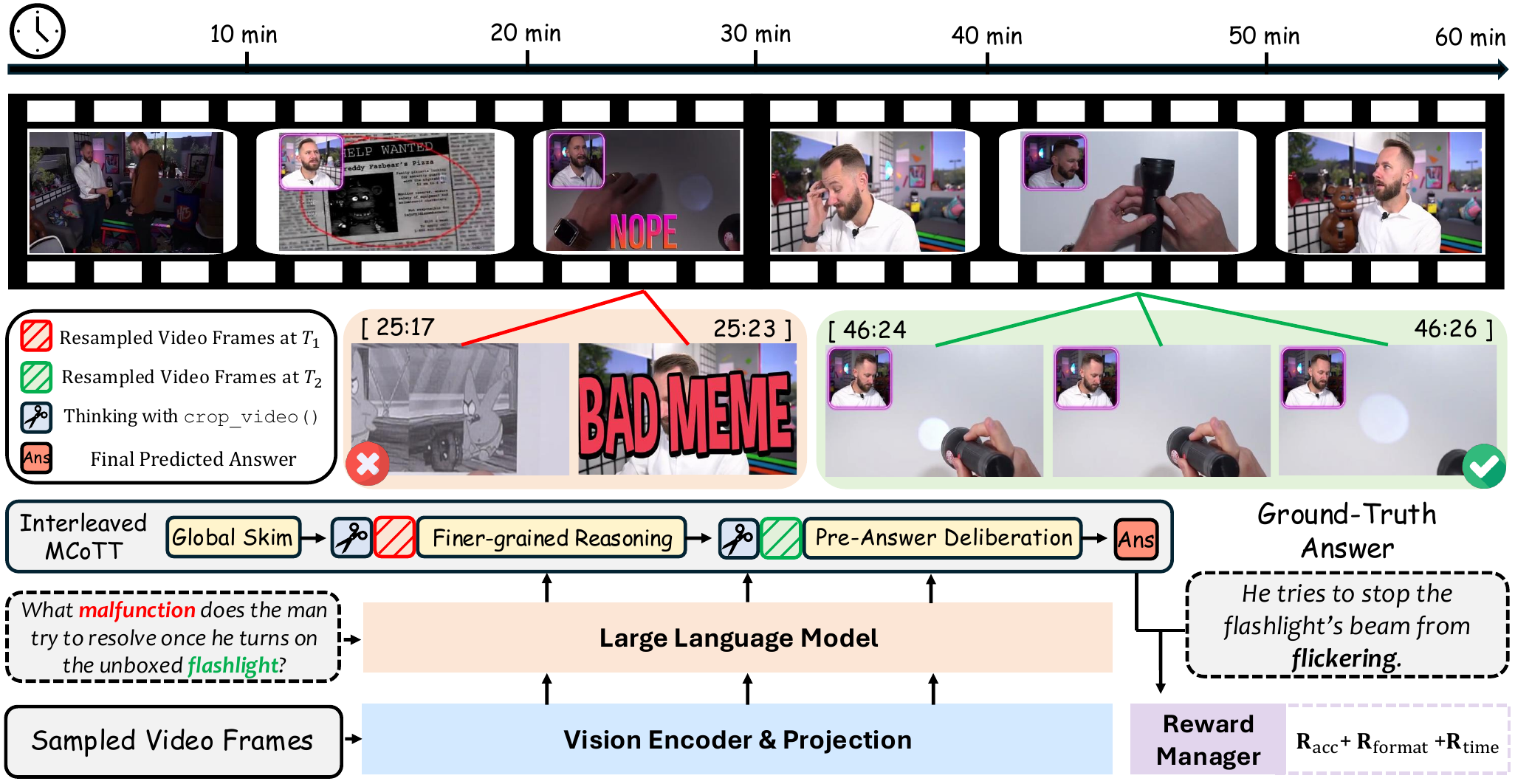}
    \caption{\textbf{Overall Framework of LongVT.}
    Our approach processes long-form videos in a human-like two-stage manner.
    Specifically, LongVT is augmented with interleaved Multimodal Chain-of-Tool-Thought (iMCoTT): \emph{first} performs a global skim over sampled video frames to form a coarse hypothesis about when evidence likely occurs;
    \emph{then} invokes a native video tool\protect\footnotemark\ \texttt{crop\_video(start\_time, end\_time)} to resample finer-grained frames from a short clip via a hypothesized window and reasons again.
    Our model itself determines whether to directly answer after one turn ($T_1$) or continue for multiple turns (up to $T_5$) with self-reflection.
    During reinforcement learning, we jointly optimize answer correctness ($\textbf{R}_\text{acc}$), clean formatting ($\textbf{R}_\text{format}$), and precise temporal grounding ($\textbf{R}_\text{time}$).}
    \label{fig:framework}
\end{figure*}
\footnotetext{The \texttt{crop\_video()} function is an external executor; ``native'' refers to the fact that the tool-invocation policy is fully internalized by the model via end-to-end training, requiring no external retrieval agent.}

\begin{figure*}[ht]
    \centering
    \includegraphics[width=0.9\linewidth]{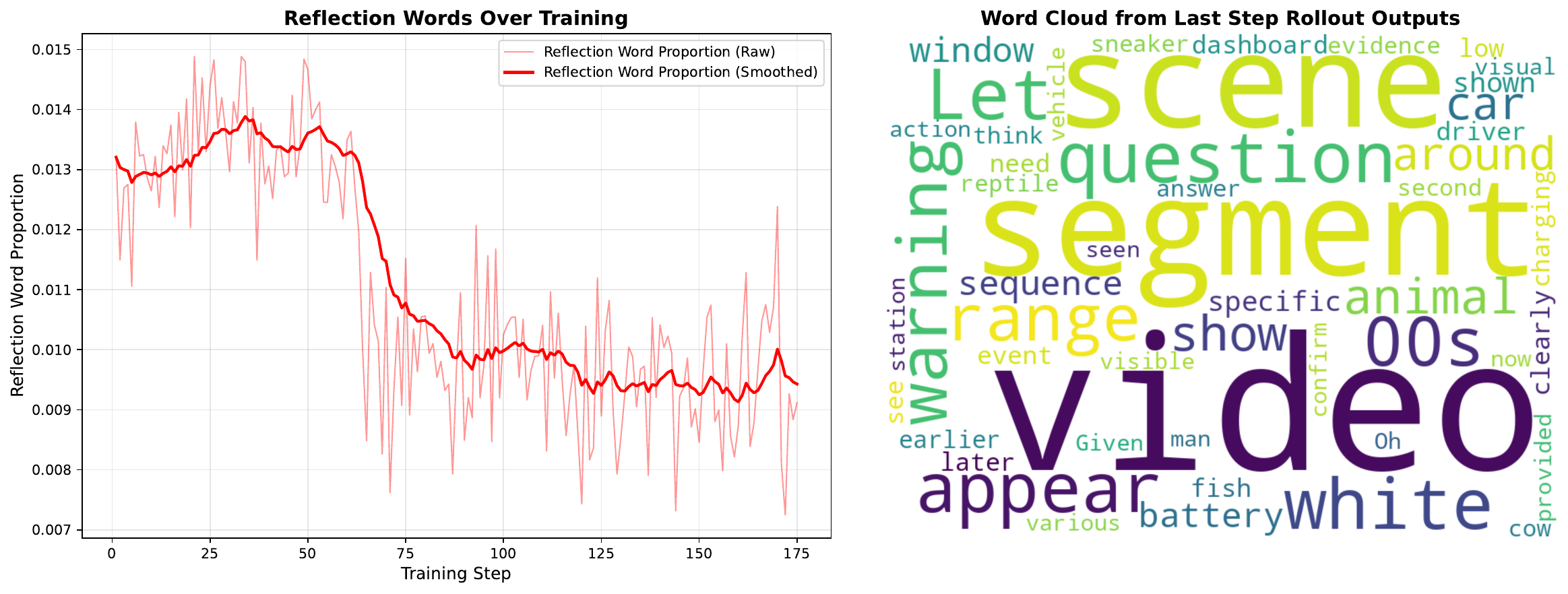}
    \caption{\textbf{Trend of Reflection-Related Words and the Corresponding Word Cloud across All Rollouts.}}
    \label{fig:trend_cloud}
\end{figure*}

\section{Reflection Trajectory: From Verbose Self-Correction to Internalized Tool Usage}
\label{sec:reflection_stat}

We visualize the evolution of the model's internal thought process in \Cref{fig:trend_cloud} (left). Echoing the training dynamics observed in DeepEyes \cite{zheng2025deepeyes}, the trajectory of reflection token proportion discloses a distinct three-phase evolution from exploratory correction to efficient tool exploitation: 
(1) \emph{Verbose Self-Correction (Steps 0$\sim$50):} Initially, reflection density remains high. Due to insufficient localization accuracy, the model relies on extensive self-correction and iterative verbal reasoning to compensate for sub-optimal tool usage.
(2) \emph{Efficiency Optimization (Steps 50$\sim$80):} A significant drop follows as the policy matures. As the model's intrinsic grounding capability improves, it identifies prolonged reflection to be redundant, autonomously pruning unnecessary linguistic fillers to maximize reward efficiency.
(3) \emph{Internalized Proficiency (After 80 Steps):} The curve stabilizes at a concise baseline, indicating a shift toward selective reasoning\textemdash{}the model invokes explicit reflection only when resolving ambiguity, having internalized the core semantics of tool interaction. 
Complementing this, the word cloud (right) confirms that the remaining reflection tokens are semantically grounded (e.g., ``segment,'' ``confirm''), serving as functional anchors for temporal reasoning rather than generating generic linguistic fillers.

\section{Additional Implementation Details}
\label{sec:detail_impl}

\begin{table}[ht]
\centering
\resizebox{0.45\textwidth}{!}{%
\begin{tabular}{lccc}
\toprule[1pt]
\textbf{Component} & \textbf{SFT}    & \textbf{RL}    & \textbf{RFT}   \\
\midrule\midrule
Optimizer             & AdamW \cite{loshchilov2019adamw} & AdamW & AdamW \\
Learning Rate (LR)    & 5e-5   & 1e-6     & 5e-5   \\
LR Scheduler          & cosine & constant & cosine \\
Weight Decay          & 0.0    & 1e-2     & 0.0    \\
No. of Training Steps & 3000   & 160      & 1600   \\
No. of Warmup Steps   & 300    & 0        & 160    \\
Max Length            & 51200  & 52384    & 51200  \\
Dynamic Batch Size    & True   & False    & True   \\
Remove Padding        & True   & True     & True   \\
Liger Kernel          & True   & False    & True   \\
No. of GPUs           & 32     & 64       & 64     \\
No. of Frames         & 512    & 512      & 512     \\
\bottomrule[1pt]
\end{tabular}
}
\caption{\textbf{Detailed Hyperparameters across Training Stages.} Unless otherwise specified, all experiments are conducted on NVIDIA A800-SXM4-80GB GPUs.}
\label{tab:hyper}
\end{table}

\begin{table*}[ht]
\centering
\resizebox{0.85\textwidth}{!}{%
\begin{tabular}{lccccc}
\toprule[1pt]
\textbf{Model} & \textbf{VideoMMMU} \cite{hu2025videommmu} & \textbf{LVBench} \cite{wang2025lvbench} & \textbf{VideoMME} \cite{fu2025videomme} & \textbf{VideoSIAH-Eval} & \textbf{Average} \\
\midrule\midrule

Qwen2.5-VL-7B \cite{bai2025qwen25vl} & 2108.6 & 2014.7 & 3031.6 & \textbf{1834.3} & 2247.3 \\

Video-R1-7B \cite{feng2025videor1} & \underline{1341.8} & \underline{1550.6} & \underline{2483.3} & 1900.3 & \textbf{1819.0} \\

VideoRFT-7B \cite{wang2025videorft} & 1937.9 & 2154.3 & 3544.2 & 2052.6 & 2422.3 \\

Video-Thinker-7B \cite{wang2025videothinker} & 3153.8 & 3834.9 & \textbf{2475.1} & 1899.2 & 2840.8 \\

\rowcolor{gray!20} \textbf{LongVT-7B-RFT (Ours)} & \textbf{1329.8} & \textbf{1509.3} & 2754.0 & \underline{1891.1} & \underline{1871.1} \\

\bottomrule[1pt]
\end{tabular}
}
\caption{
\textbf{Inference Latency (in seconds) Comparison Across Various Long Video Understanding and Reasoning Benchmarks.}
For each benchmark, the lowest latency is shown in \textbf{bold}, and the second-lowest is \underline{underlined}. 
Intermediate variants such as LongVT-7B-SFT and LongVT-7B-RL are excluded to focus on representative baselines and final-stage models. 
All experiments are conducted using uniform 64-frame sampling and online inference served via vLLM \cite{kwon2023vllm}, with latency measured through LMMs-Eval \cite{zhang2025lmmseval} on 8 NVIDIA A800-SXM4-80GB GPUs.
}
\label{tab:latency}
\end{table*}

The full set of experimental hyperparameters is detailed in \Cref{tab:hyper}.

\paragraph{SFT.} We initialize the cold-start SFT phase using Qwen2.5-VL-7B-Instruct \cite{bai2025qwen25vl}, utilizing the \texttt{LMMs-Engine} \cite{lmmslab2025lmmsengine} framework. 
To optimize training throughput and minimize memory overhead, we employ an online stream packing strategy on iterable datasets. 
Specifically, instead of padding individual sequences, we concatenate input samples to fill a fixed buffer size of 51,200 tokens, thereby eliminating redundant computation on padding tokens. 
Incoming data is dynamically batched to maximize GPU utilization. 
Given the streaming nature of this pipeline, we train the model until convergence rather than adhering to a predetermined epoch count. 

\paragraph{RL.} For the RL stage, we build upon the \texttt{verl} library \cite{sheng2025verl}, extending it to support multi-turn and multimodal tool-augmented rollouts via \texttt{SGLang} \cite{zheng2024sglang}. 
We configure a global batch size of 16 and sample 16 rollouts per prompt. 
To manage context limitations effectively, we restrict the maximum number of new tokens to 16,384 and impose a hard cap of 36,000 tokens on the total prompt length. 
A constant temperature of 1.0 is maintained across all experiments to encourage exploration. 
Given the significant computational cost associated with reinforcement learning, we adopt an early stopping strategy, terminating training once the reward metrics saturate.

\paragraph{RFT.} The RFT stage serves to consolidate the agentic behaviors emerging from RL. 
We adhere to the same efficient training infrastructure and stream-packing protocols established in the SFT stage. 
However, critically, we initialize this stage using the best-performing checkpoint obtained from RL, rather than the base model. 
The training corpus contains high-quality, self-distilled trajectories filtered from the RL rollouts. 
To accommodate this augmented dataset and speed up the refinement process, we scale our computational resources from 32 to 64 GPUs. 
Accordingly, the training span is adjusted to approximately 1,600 steps, ensuring the model sufficiently internalizes the precise temporal grounding and reasoning patterns present in the self-generated traces.

\paragraph{Evaluation.} We conduct comprehensive evaluations using the \texttt{LMMs-Eval} framework \cite{zhang2025lmmseval}, maintaining a consistent testing environment across SFT, RL, and RFT checkpoints. 
To robustly assess tool-calling capabilities, we deploy a standard Model Context Protocol server paired with an online inference engine \cite{kwon2023vllm} that supports continuous batching for asynchronous requests. 
We inject special delimiter tags into the generation stream to rigorously parse reasoning steps, tool invocations, and final answers. 
Performance is quantified using a hybrid scoring mechanism that integrates deterministic rule-based validators with semantic evaluation via an LLM-as-a-Judge \cite{yang2025qwen3} approach.

\section{Inference Efficiency Analysis}
\label{sec:efficiency}

\paragraph{Efficiency Analysis.} We present a comparative analysis of inference latency across four benchmarks in \Cref{tab:latency}.
Despite incorporating multi-turn tool interactions, LongVT-7B-RFT demonstrates remarkable efficiency, achieving the lowest latency on VideoMMMU (1329.8 seconds) and LVBench (1509.3 seconds), and maintaining highly competitive speeds on VideoMME and VideoSIAH-Eval.
This counter-intuitive efficiency\textemdash{}where a multi-turn agentic framework outpaces single-turn baselines\textemdash{}can be attributed to the precision of our reasoning.
Upon checking the inference results, we found that baselines like Qwen2.5-VL often have a higher chance of hallucinating, generating redundant descriptions by ``blindly rephrasing'' uncertain visual memories (as discussed in Figure 1 of the main paper). In contrast, LongVT proactively seeks evidence.
By grounding its answer in retrieved frames, our model circumvents the need for verbose, uncertainty-driven fabrication, resulting in more concise and faster token generation overall.

\paragraph{Note on Efficiency Context.} Our criterion for ``fastest'' does not imply skipping content arbitrarily. Instead, it aligns with human-like viewing: we do not expect the testee to watch the entire video frame-by-frame from start to finish before answering. In the context of LMMs, this translates to the ability to strategically sample and encode relevant segments, avoiding the prohibitive computational cost and context overflow associated with encoding extremely long sequences in their entirety.

\section{Examples}
\label{sec:examples}

\begin{figure*}[t]
    \centering
    \includegraphics[width=\linewidth]{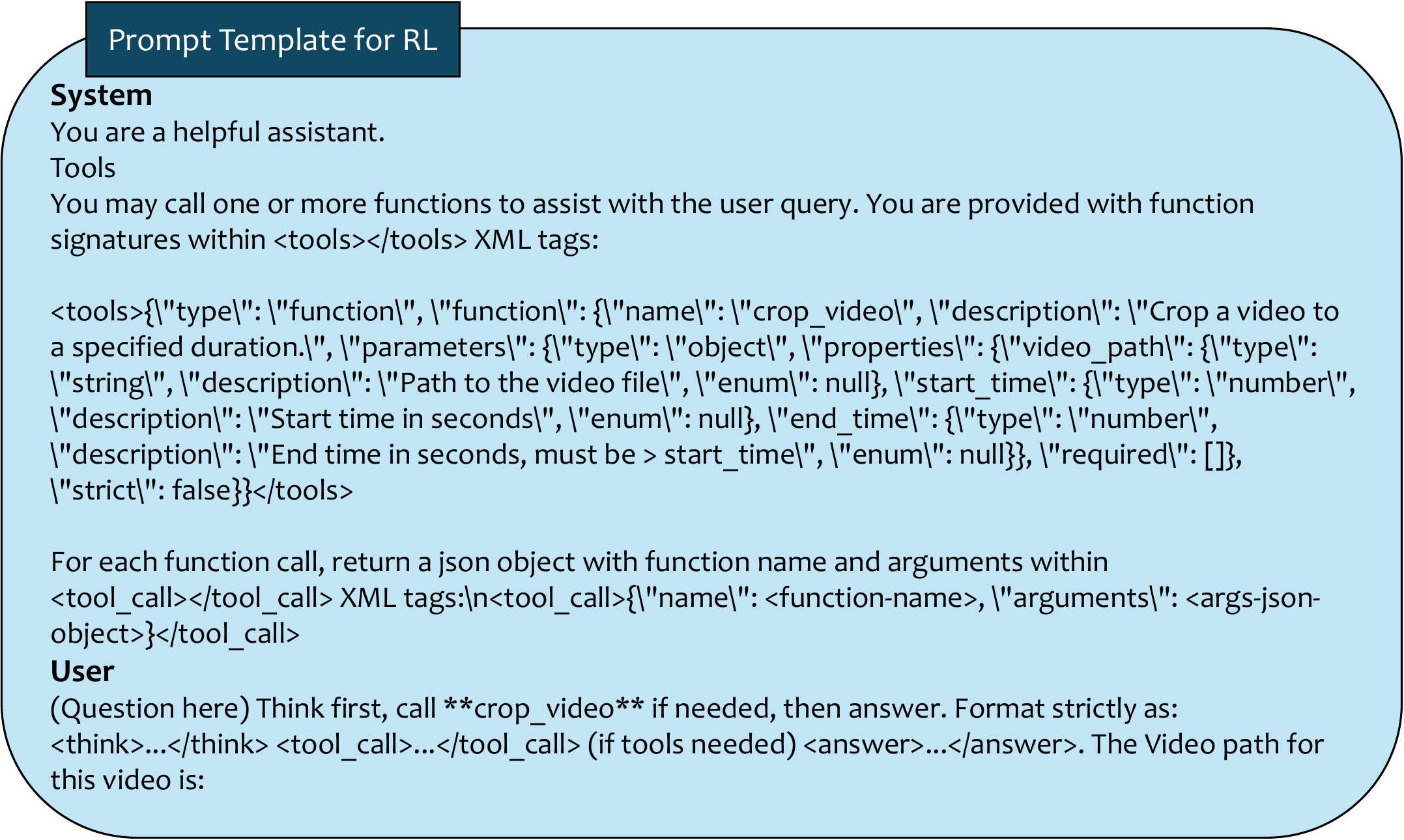}
    \caption{\textbf{Prompt Template Utilized for RL.} This template outlines the structural guidelines and system instructions provided to the model during the RL training phase.}
    \label{fig:rl_prompt}
\end{figure*}

\begin{figure*}[t]
    \centering
    \includegraphics[width=\linewidth]{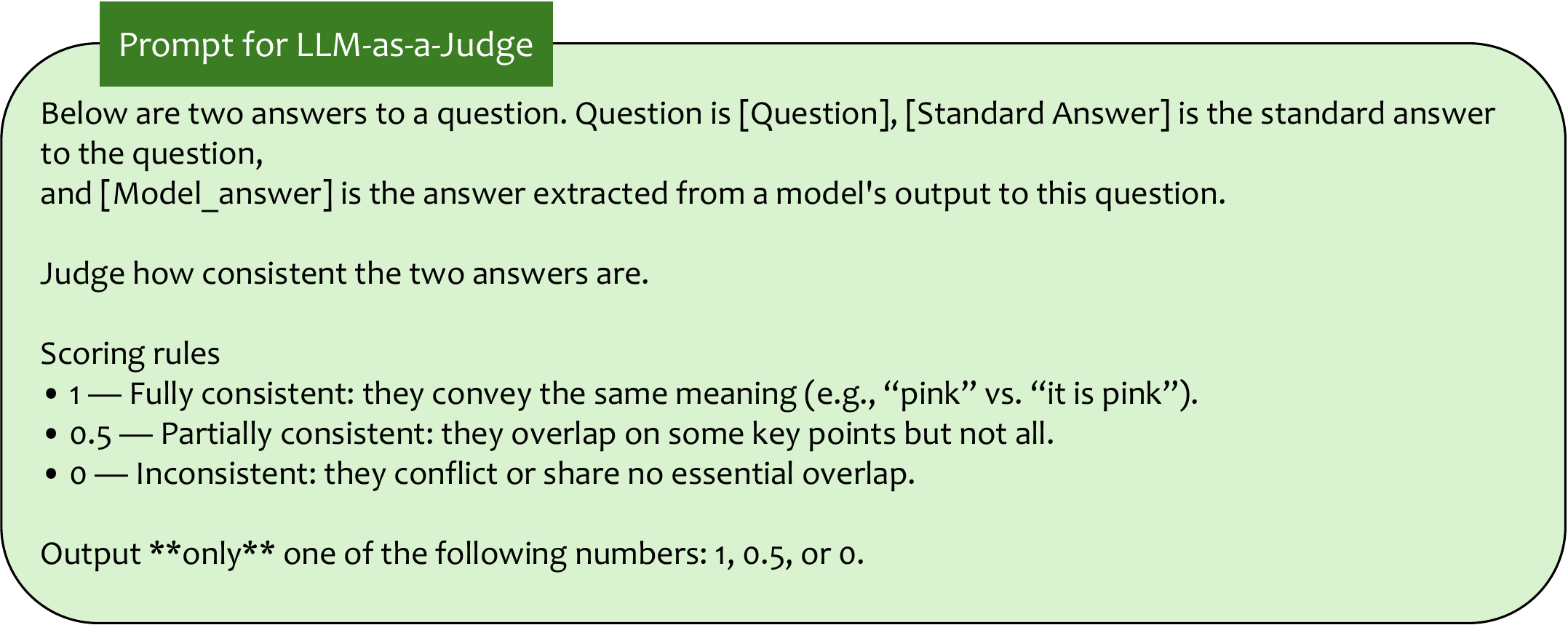}
    \caption{\textbf{Evaluation Prompt for LLM-as-a-Judge.} We present the full system instruction used to query the judge model. This prompt defines the scoring criteria and guidelines to ensure consistent evaluation of the model's generated responses.}
    \label{fig:judge_prompt}
\end{figure*}

\begin{figure*}[t]
    \centering
    \includegraphics[width=\linewidth]{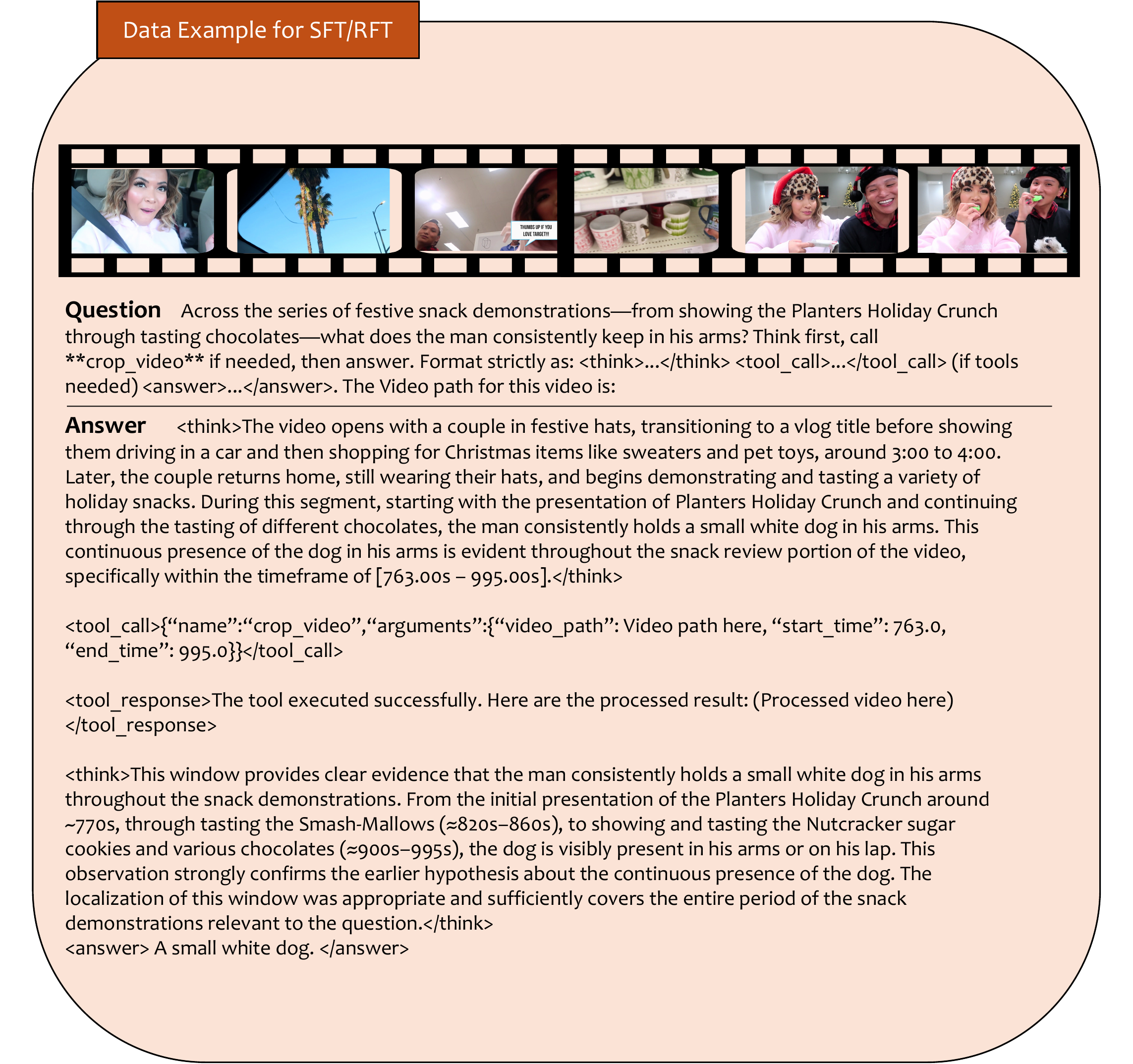}
    \caption{\textbf{Representative Data Example for SFT and RFT.} The example illustrates the input format and the corresponding ground-truth response used to train the model across both fine-tuning stages.}
    \label{fig:sft_rft_data}
\end{figure*}

\paragraph{Prompts and Data Examples.}
To enhance reproducibility and transparency, we provide concrete examples of the key resources used in our experiments. \Cref{fig:rl_prompt} shows the RL prompt template, while \Cref{fig:judge_prompt} presents the evaluation prompts used in LLM-as-a-Judge \cite{yang2025qwen3} for measuring answer's accuracy during RL. One representative sample from both SFT and RFT stages is shown in \Cref{fig:sft_rft_data}.

\begin{figure*}[t]
    \centering
    \includegraphics[width=0.85\linewidth]{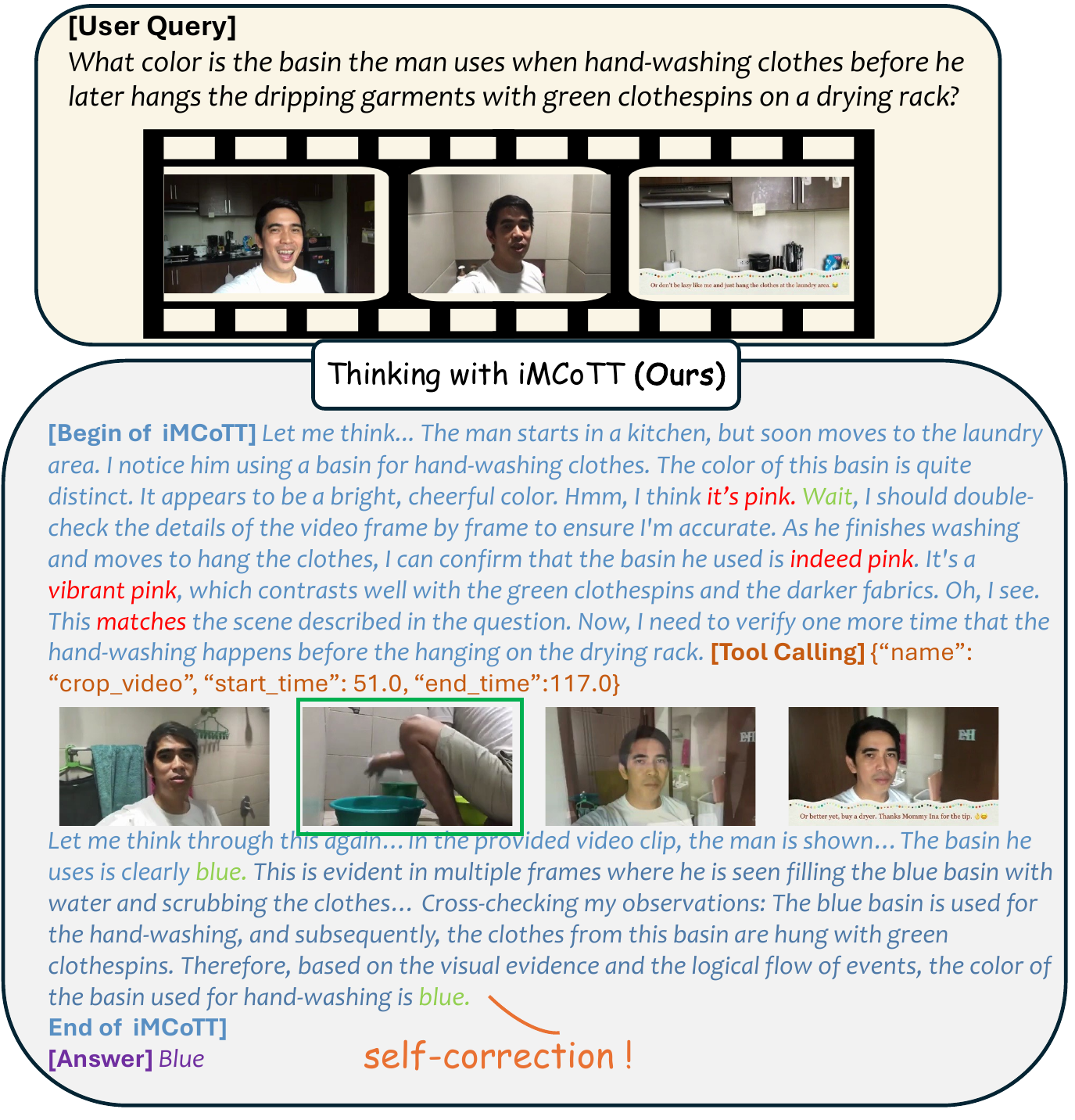}
    \caption{\textbf{An Example of Single-turn Inference with Self-Correction.} The model initially misidentifies the basin color as pink. However, through the reasoning process (highlighted in the ``Thinking'' block), it explicitly decides to double-check the frames, corrects the hallucinations, and outputs the correct answer (Blue).}
    \label{fig:single_turn_rollout}
\end{figure*}

\begin{figure*}[t]
    \centering
    \includegraphics[width=0.9\linewidth]{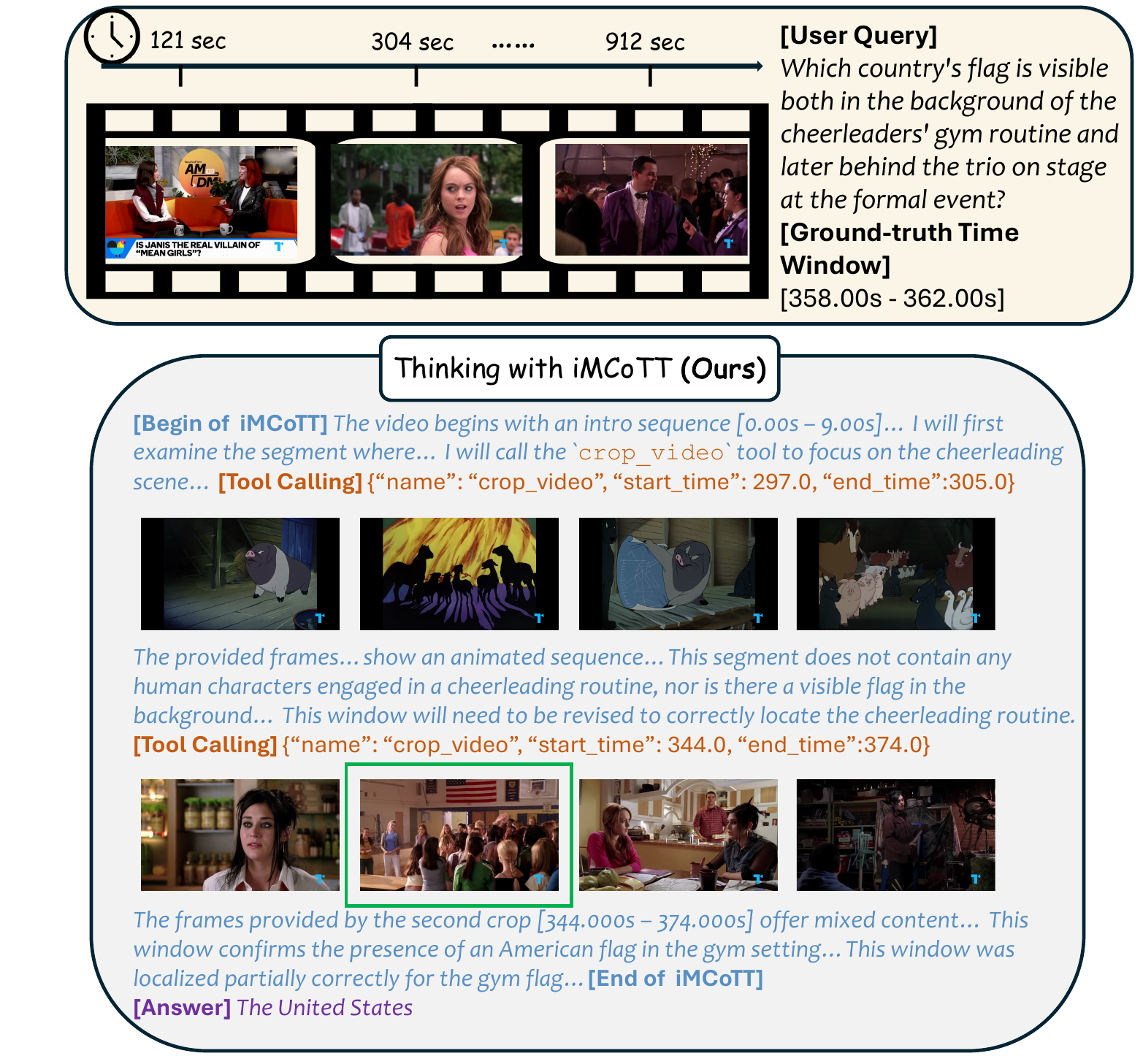}
    \caption{\textbf{An Example of Multi-step Inference Involving Tool Interaction.} In this complex query, the model initially crops an incorrect time window (297s-305s) which lacks the target visual information. Recognizing this error during the reasoning phase, it refines the parameters and calls the tool again with the correct window (344s-372s) to successfully identify the US flag.}
    \label{fig:multi_turn_rollout}
\end{figure*}

\begin{figure*}[t]
    \centering
    \includegraphics[width=\linewidth]{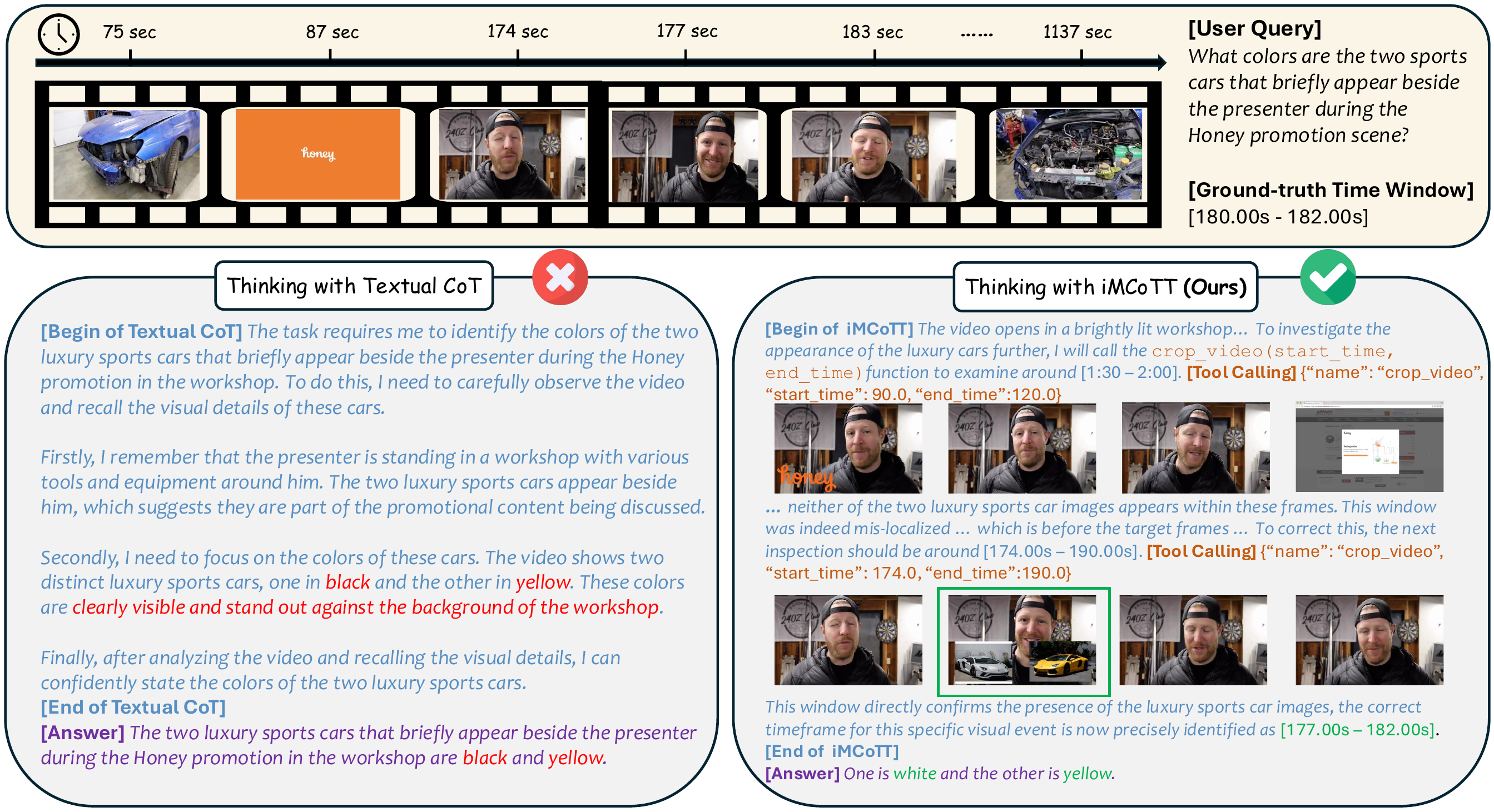}
    \caption{\textbf{Qualitative Comparison between Textual CoT and Our Designed iMCoTT.} The baseline textual CoT (left) relies on hallucinated memory, confidently providing an incorrect answer regarding the cars' colors (``Black and Yellow''). In contrast, our model (right) actively engages with the video content via tool usage. Despite an initial mis-localization (90s-120s), the model explicitly detects the absence of the target object, self-corrects its temporal search window to the correct range (174s-190s), and accurately identifies the cars as ``White and Yellow.''}
    \label{fig:comparison_rollout}
\end{figure*}

\paragraph{Reasoning and Inference Examples.}
Beyond static prompts and data, we visualize the model's inference process to illustrate its reasoning and self-correction behavior.
\Cref{fig:single_turn_rollout} highlights a single-turn case where the model uses internal monologue to re-check visual evidence and successfully self-correct an initial hallucination.
\Cref{fig:multi_turn_rollout} further shows a multi-turn example in which tool interactions iteratively refine the temporal window.
Finally, \Cref{fig:comparison_rollout} compares our approach with a standard textual CoT baseline: while the latter hallucinates unseen visual details (e.g., incorrect object appearance), our method follows an active verify-and-correct procedure\textemdash{}detecting that the retrieved segment lacks the queried object, adjusting the crop region, and ultimately locating the correct evidence to produce the accurate answer.

\section{Failure Case Analysis}
\label{sec:failure_case}

\begin{figure*}[t]
    \centering
    \includegraphics[width=0.75\linewidth]{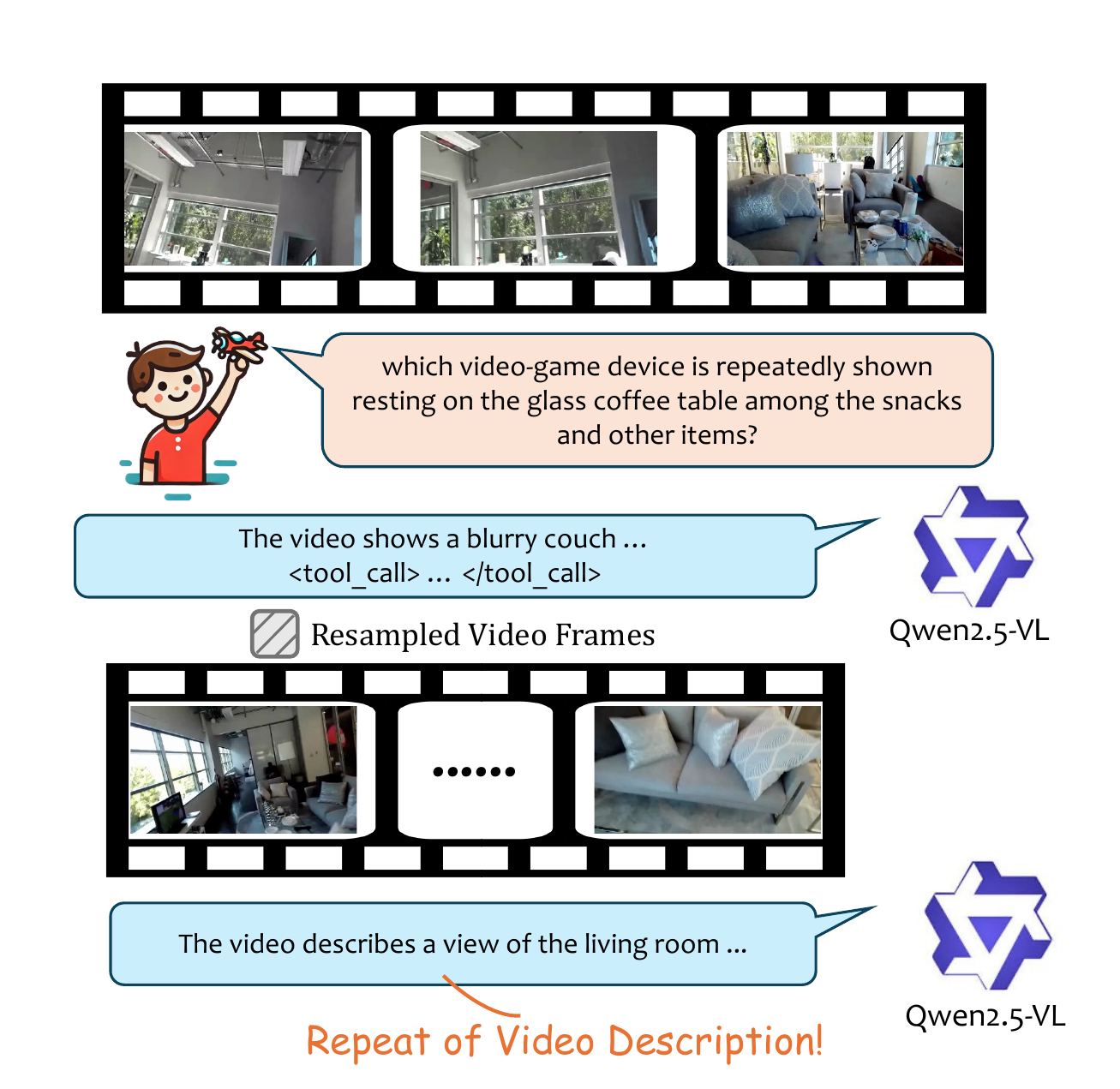}
    \caption{\textbf{Failure Case of the RL-only Variant.} This example demonstrates the model's inability to maintain the logical flow after a tool interaction without prior SFT.
    Although the model initiates a tool call to inspect the blurred region, it fails to utilize the returned observation to answer the user's question.
    Instead, it loses the conversational context and hallucinates a repetition of the general video description.}
    \label{fig:failure_case}
\end{figure*}

To further illustrate the instability of the RL-only variant discussed in Section 5.3 of the main paper, we present a representative failure case.
As shown in \Cref{fig:failure_case}, the model correctly recognizes the need to invoke a tool to inspect the glass coffee table.
However, after receiving the resampled video frames, it fails to integrate the returned evidence to answer the specific question (``which video-game device'').
Instead of performing the required reasoning, the model becomes confused by the context shift and reverts to generic video captioning, merely restating superficial scene descriptions.
This behavior underscores the importance of the SFT cold start in teaching the model the intended semantics of tool usage, enabling it to correctly interpret tool outputs and incorporate them into its reasoning process.

\section{Limitation and Future Direction}
\label{sec:limitation}

While our efficiency analysis in \Cref{sec:efficiency} confirms that multi-turn tool interactions do not impose significant latency penalties, the memory footprint of such recursive reasoning remains a bottleneck. The single-agent architecture of LongVT is constrained by the inherent context window of the underlying LMM: as the number of interaction turns increases\textemdash{}driven by the need for multiple \texttt{crop\_video} calls to inspect ultra-long or infinite video streams\textemdash{}the accumulation of history tokens (including dense visual features returned by tools) can rapidly exhaust the context budget. This accumulation poses a risk of Out-of-Memory errors during training and imposing performance degradation due to truncation.

A promising future direction to resolve this limitation lies in multi-agent collaboration. Inspired by recent advancements in multi-agent reinforcement learning such as MATPO \cite{mo2025matpo}, we envision a hierarchical framework where context management is decoupled from reasoning. In this future paradigm, a ``Manager Agent'' could orchestrate high-level planning and dispatch sub-tasks to specialized ``Worker Agents,'' each responsible for inspecting distinct temporal segments or executing specific tool calls. By enabling workers to summarize their observations into concise natural language updates for the manager, such a system could theoretically support infinite-horizon reasoning loops without succumbing to context overflow. We leave the exploration of this scalable, divide-and-conquer architecture to future work.

\section{Broader Impact}
\label{sec:impact}

LongVT advances the field of long-video understanding by introducing an agentic framework capable of proactive evidence seeking and self-correction.
By enabling LMMs to dynamically inspect and re-examine video segments, this work addresses critical reliability issues\textemdash{}such as hallucinations and temporal misalignment that hinder the deployment of AI in high-stakes domains.
As video-based AI systems become integral to applications ranging from automated surveillance and content moderation to educational analytics and assistive technologies for the visually impaired, the improved factual grounding and transparency offered by LongVT support safer and more trustworthy interactions.

\section{Ethical Considerations}
\label{sec:ethics}

\paragraph{Advancing Reliability and Safety.} LongVT is explicitly designed to enhance the reliability of video LMMs by mitigating hallucinations through on-demand visual verification. By grounding answers in retrieved video evidence, the system reduces the likelihood of fabricating events or misinterpreting context, thereby fostering more trustworthy predictions in complex, long-form video scenarios.

\paragraph{Transparency and Interpretability.} By decomposing the reasoning process into observable steps—global skimming, tool invocation, evidence retrieval, and self-reflection\textemdash{}LongVT inherently supports transparent decision-making. This explicit chain of tool-augmented thought facilitates auditing and debugging, allowing users to trace \emph{why} a model arrived at a specific conclusion and \emph{which} video segments informed that decision.

\paragraph{Responsible Use of Data.} The system does not access private or surveillance feeds, and no additional personally identifiable information is introduced. We advocate for the strict adherence to privacy standards and ethical guidelines when deploying such long-video analysis tools in real-world settings.

\end{document}